\pdfoutput=1

\documentclass[10pt,twocolumn]{article}
\usepackage{times}
\usepackage{fullpage}
\usepackage{color}
\usepackage{anyfontsize}
\usepackage{ifthen}
\usepackage{xspace}
\usepackage{amsmath}
\usepackage[compact,small]{titlesec}
\usepackage{listings}
\usepackage{amssymb}
\usepackage{url}
\usepackage{graphicx}
\usepackage{soul}
\usepackage{subfig}
\usepackage{enumitem}
\usepackage{tabularx}
\usepackage[T1]{fontenc}
\usepackage[left=0.75in,right=0.75in,top=1in,bottom=1in]{geometry}

\pdfpagewidth=\paperwidth
\pdfpageheight=\paperheight
\titlespacing*{\section}{0pt}{3pt}{3pt}
\titlespacing*{\subsection}{0pt}{3pt}{3pt}
\titlespacing*{\subsubsection}{0pt}{3pt}{3pt}

\graphicspath{{figs/}{data/}}

\setlist[enumerate]{leftmargin=0pt,labelindent=6pt, labelwidth=!}

\newboolean{comments}
\setboolean{comments}{false}
\newboolean{anon}
\setboolean{anon}{false}

\newcommand{\ie}{{\em i.e.,~\xspace}}
\newcommand{\eg}{{\em e.g.,~\xspace}}
\newcommand{\Eg}{{\em E.g.,~\xspace}}
\newcommand{\smallurl}[1]{{\scriptsize \url{#1}}}

\newcommand{\reffig}[1]{Fig.~\ref{fig:#1}}

\newcommand{\refsec}[1]{\S\ref{sec:#1}}
\newcommand{\refeq}[1]{Eq.~\ref{eq:#1}}
\newcommand{\reftbl}[1]{Table~\ref{tbl:#1}}

\newcommand{\eps}{\epsilon}

\newcommand{\algofont}[1]{\ensuremath{\mathtt{#1}}}
\newcommand{\mwt}{\ensuremath{\mathtt{MWT}}\xspace}
\newcommand{\MWT}{\ensuremath{\mathtt{MWT}}\xspace}
\ifthenelse{\boolean{anon}} {
\newcommand{\ds}{Anon Service\xspace}
\newcommand{\DS}{Anon Service\xspace}
\newcommand{\msn}{XYZ\xspace}
\newcommand{\tr}{GHI\xspace}
\newcommand{\ac}{MNO\xspace}
\newcommand{\complex}{DEF\xspace}
\newcommand{\toronto}{TSA\xspace}
} {
\newcommand{\ds}{Decision Service\xspace}
\newcommand{\DS}{Decision Service\xspace}
\newcommand{\msn}{MSN\xspace}
\newcommand{\tr}{TrackRevenue\xspace}
\newcommand{\ac}{Azure Compute\xspace}
\newcommand{\complex}{Complex\xspace}
\newcommand{\toronto}{Toronto\xspace}
}
\newcommand{\news}{\ensuremath{\mathtt{News}}\xspace}

\newcommand{\job}{\ensuremath{\mathtt{Cloud}}\xspace}
\newcommand{\techsupport}{\ensuremath{\mathtt{TechSupp}}\xspace}
\newcommand{\EpsGreedy}{\algofont{EpsilonGreedy}\xspace}
\newcommand{\muIPS}{\mathtt{ips}}
\newcommand{\ips}{\ensuremath{\mathtt{ips}}\xspace}
\newcommand{\CSC}{cost-sensitive classification\xspace}

\newcommand{\vw}{\algofont{VW}\xspace}

\usepackage{dsfont} 
\newcommand{\indicator}[1]{\mathds{1}{\{#1\}}}
\newcommand{\APP}{\ensuremath{\mathtt{APP}}\xspace}
\newcommand{\about}[1]{$\sim${#1}}

\newcommand{\rulesep}{\unskip\ \vrule\ }

\newenvironment{noindentlist2}
 {\begin{list}{\labelitemi}{\leftmargin=0pt \itemindent=1em \itemsep=1pt}}
 {\end{list}}

\ifthenelse{\boolean{comments}} {
  \usepackage[draft]{todonotes}
  \setlength{\marginparwidth}{2cm}

  \newcommand{\comment}[1]{\textcolor{red}{(#1)}}
  \newcommand{\john}[1]{\textcolor{red}{(John: #1)}}
  \newcommand{\sid}[1]{\textcolor{green}{(Sid: #1)}}
  \newcommand{\alex}[1]{\textcolor{magenta}{(Alex: #1)}}
  \newcommand{\alekh}[1]{\textcolor{olive}{{Alekh: #1}}}
  \newcommand{\gal}[1]{\textcolor{blue}{{Gal: #1}}}
  \newcommand{\markus}[1]{\textcolor{orange}{{Markus: #1}}}
} {
  \usepackage[disable]{todonotes} 
  \newcommand{\comment}[1]{}
  \newcommand{\john}[1]{}
  \newcommand{\sid}[1]{}
  \newcommand{\alex}[1]{}
  \newcommand{\alekh}[1]{}
  \newcommand{\gal}[1]{}
  \newcommand{\markus}[1]{}
}
\newcommand{\ignore}[1]{}
\newcommand{\msaffil}{$^*$}

\usepackage[suppress]{color-edits}  
\addauthor{as}{red}     
\addauthor{ss}{blue}    
\addauthor{sb}{magenta}   
\addauthor{aa}{olive}   

\newcommand{\OMIT}[1]{}
\newcommand{\xhdr}[1]{\vspace{0mm} \noindent{\bf #1}}

\let\oldenumerate\enumerate
\renewcommand{\enumerate}{
  \oldenumerate
  \setlength{\itemsep}{.5pt}
  \setlength{\parskip}{0pt}
  \setlength{\parsep}{0pt}
  \setlength{\topsep}{0pt}
}

\let\olditemize\itemize
\renewcommand{\itemize}{
  \olditemize
  \setlength{\itemsep}{1pt}
  \setlength{\parskip}{0pt}
  \setlength{\parsep}{0pt}
  \setlength{\topsep}{0pt}
}

\definecolor{bluekeywords}{rgb}{0,0,1}
\definecolor{greencomments}{rgb}{0,0.5,0}
\definecolor{redstrings}{rgb}{0.64,0.08,0.08}
\definecolor{types}{rgb}{0.17,0.57,0.68}
\definecolor{identifiers}{RGB}{43,145,175}

\begin{document}
\title{\vspace{-30pt}\bf Making Contextual Decisions
with Low Technical Debt\vspace{-1ex}}

\author{Alekh Agarwal, Sarah Bird, Markus Cozowicz, 
Luong Hoang, John Langford, Stephen Lee\msaffil, Jiaji Li\msaffil
\and Dan Melamed, Gal Oshri\msaffil, Oswaldo Ribas\msaffil,
 Siddhartha Sen, Alex Slivkins \\[-1pt]
\centerline{\small{Microsoft Research, \msaffil Microsoft}}
}
\date{\vspace{-22pt}}
\maketitle
\thispagestyle{empty}

\setlength{\headsep}{0.2in}


\begin{abstract}
\vspace{0pt}
Applications and systems are constantly faced with decisions that require
picking from a set of actions based on contextual information.
Reinforcement-based learning algorithms such as contextual bandits can be very
effective in these settings, but applying them in practice is fraught with
technical debt, and no general system exists that supports them completely. We
address this and create the first general system for contextual learning, called
the \ds.

Existing systems often suffer from technical debt that arises from
issues like incorrect data collection and weak debuggability, issues
we systematically address through our ML methodology and system
abstractions. The \ds enables all aspects of contextual bandit
learning using four system abstractions which connect together in a
loop: explore (the decision space), log, learn, and deploy. Notably,
our new explore and log abstractions ensure the system produces
correct, unbiased data, which our learner uses for online learning and
to enable real-time safeguards, all in a fully reproducible manner.

The \ds has a simple user interface and works with a variety of
applications: we present two live production deployments for content
recommendation that achieved click-through improvements of 25-30\%,
another with $18\%$ revenue lift in the landing page, and
ongoing applications in tech support and machine failure handling. The
service makes real-time decisions and learns continuously and
scalably, while significantly lowering technical debt.  

\ignore{
Applications and systems are constantly faced with decisions to make that
must pick from a set of actions based on some contextual information. Contextual
bandits is a type of machine learning for these situations, but applying it in
practice is fraught with technical debt, and no general system exists that
supports it. We address this and create the first general system for contextual
learning, called the \ds.

The technical debts we address, which arise from things like incorrect data
collection and weak debuggability, guide our use of ML techniques and the design
of our system. In particular, we synthesize an ML methodology and then realize
it with the help of four system abstractions connected in a loop to:
explore (the decision space), log, learn, and deploy. Notably, our new
explore and log abstractions ensure the system produces correct, unbiased
data, which our learner uses for online learning and to enable
real-time safeguards.

 The \ds has a simple user interface and works with a variety of
applications: we report on three live production deployments for content
recommendation that saw click-throughs rise by 14--25\%, and two ongoing
applications in health and machine failure handling. We show that the service
makes decisions in real time and learns from new data quickly and scalably,
while significantly lowering technical debt.
}

\ignore{
-apps faced with contextual decisions
-there exists learning theory for this, but adapting it to practical is hard
-moreover, very easy to fall into technical debt traps, as the experience of
ours and others has shown
-we provide a service for solving our problem that addresses several major
technical debt issues, and in doing so we arrive at a robust solution to the
problem
-The technical debts we address guide our synthesis of the ML and the design of
our system. In particular, we propose several key abstaactions of such a system
that address the technical debt issues: these include exploration, join service,
and learning. 
-In turn, these abstractions give rise to a very general system with a
simple UI that can be used for a variety of applications. We describe several
deployments, three of them in production and two in prototyping stage, that
integrate into our system in different ways (highlighting its flexibility) 
-We evaluate the performance of our system not just by standard system
metrics but also its ability to reduce technical debt.
}

\ignore{
Applications and systems are constantly faced with decisions to make,
often using a \emph{policy} to pick from a set of actions based on
some contextual information. We create a service that uses machine
learning to accomplish this goal.  The service uses exploration,
logging, and online learning to create a counterfactually sound system
supporting a full data lifecycle.

The system is general: it works for any discrete choices, with respect
to any reward metric, and can work with many learning algorithms and
feature representations.  The service has a simple API, and was
designed to be modular and reproducible to ease deployment and
debugging, respectively.  We demonstrate how these properties enable
learning systems that are robust and safe.

Our evaluation shows that the \ds makes decisions in real time and
incorporates new data quickly into learned policies.  A large-scale
deployment for a personalized news website has been handling all
traffic since Jan.  2016, resulting in a 25\% relative lift in clicks.
By making the \ds externally available, we hope to make optimal
decision making available to all.
}

\end{abstract}

\section{Introduction}
\label{sec:intro}

Machine learning has well-known high-value applications, yet effective
deployment is fraught with difficulties in
practice~\cite{bottou-ad,hitd}.  Machine-learned models for
recommendation, ranking, and spam detection all become obsolete unless
they are regularly infused with new data.  Using data is actually
quite tricky---thus the growing demand for data scientists. How do we
create a machine learning system that \asedit{collects and uses data
  in a safe and correct manner?}

Recently, Sculley et al.~\cite{hitd} used the framework of {\em
  technical debt}---the long-term costs that accumulate when expedient
(but suboptimal) decisions are made in the short run---to argue that
ML systems incur massive hidden costs at the system level, beyond the
basic code complexity issues of traditional software systems. Thus,
traditional mechanisms for coping with technical debt, such as
refactoring code or improving unit tests, are insufficient for ML
systems.

In this work, we create a {\em system for contextual decision making}
that addresses an important subset of these \asedit{failure modes and}
technical debts. Prior experiences of deploying one-off
solutions~\cite{Deepak,Langford-www10,Langford-wsdm11,bottou-ad}, and
failures we have experienced ourselves, call for a systematic approach
that addresses these debts by design. Specifically, we target
\asedit{failure modes and technical debts} incurred by: feedback loops
and bias, distributed data collection, changes in the environment, and
weak monitoring and debugging (\refsec{motivation}).  Existing systems
neither solve the class of problems we handle, nor address these
debts.

As it turns out, the problem domain we are interested in---contextual
decisions in interactive settings---leads to higher incidence of
\asedit{failure modes and} technical debt than traditional supervised
prediction problems. In a traditional prediction problem, a context
(\eg image features) is given, a prediction (\eg dog or cat) is made
and the true label is then revealed. Crucially, the label can be used
to measure the quality of \emph{any potential prediction} on this
context, which is key in the supervised learning techniques used for
this setting. In our setting, an application repeatedly takes
\emph{actions} (\eg which type of news article to show) when faced
with a particular \emph{context} (\eg user demographics) to obtain a
desirable outcome quantified as a \emph{reward} (\eg a click). The
goal is to find a good \emph{policy} mapping contexts to actions, \eg
show politics articles to adults and sports articles to teenagers. The
feedback mode puts the algorithm in a reinforcement learning setting,
which is significantly more challenging~\cite{RegressorElim-aistats12}
than supervised learning~\cite{FS97}. This paradigm covers a huge
range of applications including virtually every online service;
\reftbl{apps} has examples.

\begin{figure}[t]
\centering
  \includegraphics[width=.32\columnwidth]{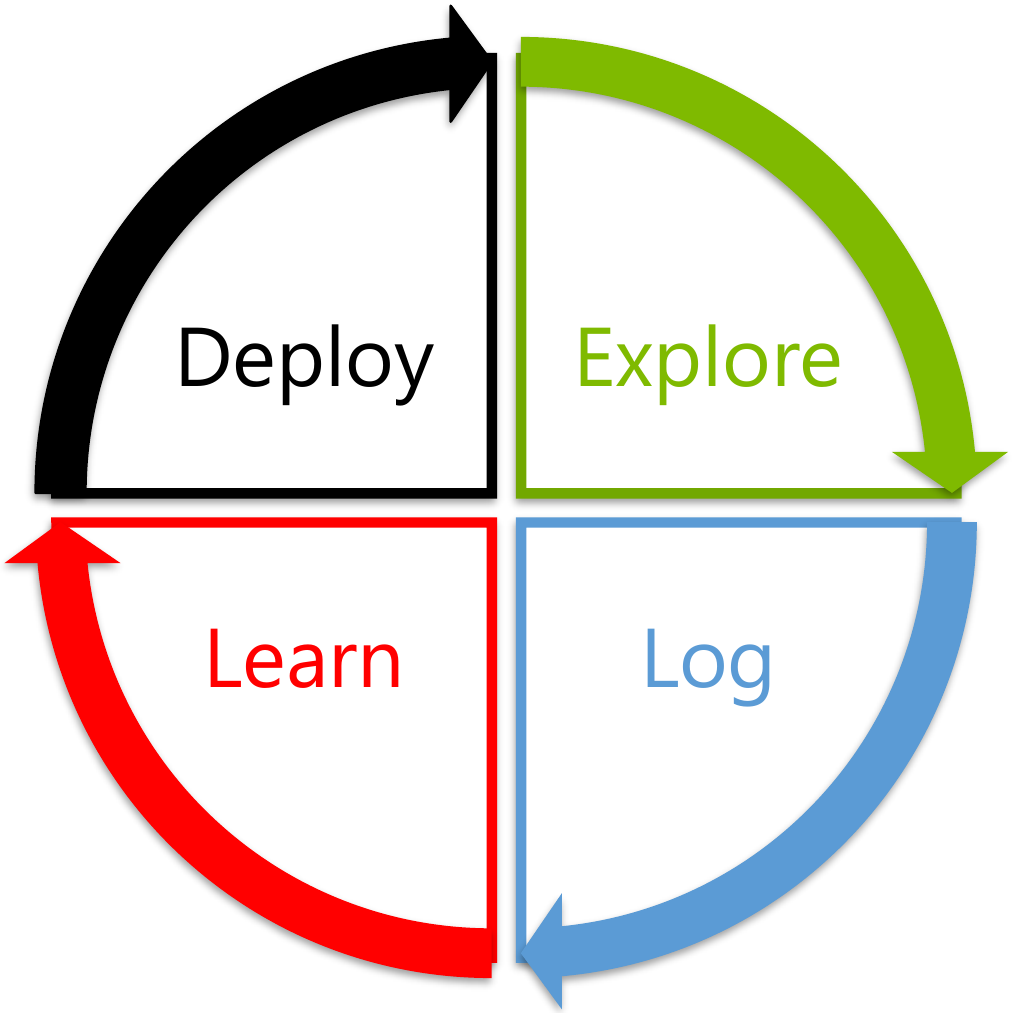}
  \caption{Complete loop for effective contextual learning, showing the four
  system abstractions we define.}
  \label{fig:cycle}
\end{figure}

\asedit{Two properties make this setting particularly challenging:}
\begin{enumerate}
\item The reward is observed only for the chosen action; nothing is
  learned about unexplored actions, leading to \emph{partial
    feedback}. In supervised learning, the true label determines the
  quality of every prediction leading to \emph{full feedback}.
\item Rewards arrive after a delay, often via a separate data path.
\end{enumerate}

\begin{noindentlist2}
\item {\bf Partial feedback.} Partial feedback settings are prone to
  bias, because the decisions made by a system affect the data
  collected to train the system~\cite{bottou-ad,hitd}.  For example,
  it is not possible to learn that a news article about sports
  generates more clicks than one about politics when shown to a
  particular demographic, without showing them both types of articles 
  at least some of the time. In addition, the
  demographic's preferences may change over time.  Hence, there is a
  need to \emph{explore} so as to acquire the right data; a
  \emph{biased} dataset, no matter how large, does not enable good
  learning.
\item {\bf Delayed rewards.} In most applications, reward information
  (\eg did the user click the article?) arrives after a considerable
  delay, ranging from seconds to days. Often, this data is collected
  by a separate subsystem and follows a separate data path.  Since an
  event is incomplete until the reward is known, this requires
  distributed data collection, which is a common source of data
  errors.  In addition, since rewards for some actions take longer to
  arrive than others, there is a real possibility of delay-related
  bias entering the dataset.
\end{noindentlist2}

Addressing these \asedit{issues} requires a synthesis of ML techniques
as well as careful system design. On the ML side, exploration
addresses some of the \asedit{issues} caused by partial feedback. A
common methodology for exploration is \emph{A/B testing}
~\cite{KohaviAB-2015,KohaviLSH09}: policy A is tested against policy B
by running both live on a percentage of user traffic drawn at random.
This requires data scaling linearly with the number of policies to be
tested. On the other hand, \emph{contextual bandits}
\cite{monster-icml14,Langford-nips07} is a type of machine learning
that allows testing and optimization over exponentially more policies
for the same amount of data\footnote{\Eg 1 billion policies for the
  data collection cost of 21 A/B tests~\cite{MWT-WhitePaper-2016}.}.
Moreover, the policies being evaluated need not be deployed or even
known ahead of time, saving vast business and engineering effort.  We
refer to this dramatic improvement in capability as {\em multiworld
  testing} (\MWT), and realize it with contextual bandits
and policy evaluation
techniques~\cite{Langford-www10,Langford-wsdm11}.

Our ML methodology has two important properties: it addresses the
biased feedback loop and enables advanced monitoring/debugging through
its policy evaluation capabilities (\refsec{mwt}). It does not,
however, \asedit{fully} address a nonstationary environment nor the data
collection issues caused be delayed rewards.  In fact, the ML theory
simply assumes correct data is provided as input, which is hard 
to ensure in practice.

We argue that well-defined system abstractions are required for
correct data collection. We propose the 4-step loop shown in
\reffig{cycle}: \emph{explore} to collect the data, \emph{log} this
data correctly, \emph{learn} a good model, and \emph{deploy} it in the
application.  Existing systems typically focus on the learn and deploy
steps, while ignoring or mishandling the data collection steps
(explore and log); further, the steps are managed by independent
processes, resulting in the complete process being inefficient,
expensive, and error-prone. By connecting these abstractions in the
right way, we address the remaining sources of \asedit{failures and}
technical debt (\refsec{arch}). The explorer invokes the logger at the
time of decision to record the decision event, and later to join the
reward. To avoid delay bias, the logger imposes a uniform delay before
releasing joined data to the learner. Finally, models are trained and
deployed online to cope with dynamic environments. The abstractions
are modular, and can be implemented, maintained and tested separately
so as to not add new sources of debt resulting from a monolithic
system.

Our system abstractions and ML methodology enable other techniques for
reducing technical debt, such as full reproducibility and real-time
safeguards (\refsec{techniques}).  The resulting system is called the
\ds: it is fully functional, publicly available, and
open-sourced~\cite{ds-clientlib}. It exposes a simple interface and
supports several deployment options (\refsec{impl}), each of which has
been exercised by a real deployment. We describe three live production
deployments for content recommendation in \msn, \complex, and \tr, and
two ongoing applications in assisting tech support staff and cloud
machine failures (\refsec{deployment}).  Each deployment has achieved
a double-digit lift between 14--25\% in their target metric (\eg
clicks, revenue) relative to strong baselines.

Our evaluation (\refsec{eval}) shows that the \ds makes decisions with
low latency, incorporates new data into deployed models quickly, and
scales to the traffic of our customers and beyond.
More importantly, we show through experiments on production data and
anecdotal experiences, that the service reduces the technical debts we sought to
address.


In summary, we make the following contributions:
\begin{noindentlist2}
\item We implement an ML methodology that achieves \mwt capability and
\asedit{eliminates important failures by design}
(\refsec{mwt}).
\item We define four system abstractions for realizing this
methodology in a robust manner (\refsec{arch}).
We describe advanced techniques that further reduce technical debt,
such as full reproducibility and real-time safeguards
(\refsec{techniques}).
\item We present the first general-purpose service for contextual reinforcement
learning, with a simple API, several deployment options, and source
code (\refsec{impl}). We describe three live production
deployments and two ongoing applications (\refsec{deployment}).
\item We evaluate the performance of the service (in production when
possible) using systems as well as learning criteria. We also evaluate its
effectiveness at reducing technical debt (\refsec{eval}).
\end{noindentlist2}

\ignore{
- Let's look at why exploration is so powerful and necessary. It enables a
capability we call MWT. We synthesize this capabiliy using existing ML. Just the use of this ML
addresses some of hte technical debt issuese we address, in particular feedback
loops and bias.
- But using existing practices to convert this to a system still leaves the
remaining technical debts. We need to carefully design our system to address
this, and in particular provide the right abstractions. We do this in our design
section, and provide abstraction for explore, logging, and learning. Decompose
the problem into 4 parts, as shown in Fig. 1; but simply identifying these
pieces and defining clear semantics is not enough. Existing systems implement
parts of them piece meal; it's the way they are connected up together that is
also important. In particular, we enforce logging at the point of decision in
the exploration library, and we join rewards with delaly in the join service,
and we learn from the data outputted by teh join service. The join service is a
simple abstraction, implementable in a variety of simple ways (we use a
streaming service, one of our customers uses a key-value store, etc.), but it is
an important and powerful abstraction because it whacks away at 2 sources of
technical debt: decoupled data colleciton and a form of bias called reward-delay
bias.
- The result of the above is the Decision Service, which we have built and
deployed, and open-sourced (mostly). To date we have completed 3 production
deployments spanning different companies, \msn.com, Complex, and
\tr. We have also started applying it to health and machine failure
diagnosis. See section X.
- Summarize our results, they've been good. Our eval focuses less on standard
system benchmarks and more on the kinds of technical debt we try to avoid.
- Contribution list
}

\ignore{
Sid's notes:
- ML has many apps, but need to safely generate and infuse new data and in
general this is hard and fraught with difficulties, leading many to talk about technical debt.
- our goal is to create a system that safely generates and uses (learns from)
new data, quickly.
- Some examplese of technical debt include decoupled data collection, poor
monitoring, etc. (Maybe explicitly list the ones we address here).
- These are exacercbated in the problem domain we consider (CB)
- we look at problem of contextual decisions, describe setup etc.
- CB raises exacerbates because: 1) need exploration but that's tricky to get
right. 2) often each part of loop managed separately, and reward timing
introduces another variable and potential source of bias

- Let's look at why exploration is so powerful and necessary. It enables a
capability we call MWT. We synthesize this capabiliy using existing ML. Just the use of this ML
addresses some of hte technical debt issuese we address, in particular feedback
loops and bias.
- But using existing practices to convert this to a system still leaves the
remaining technical debts. We need to carefully design our system to address
this, and in particular provide the right abstractions. We do this in our design
section, and provide abstraction for explore, logging, and learning. Decompose
the problem into 4 parts, as shown in Fig. 1; but simply identifying these
pieces and defining clear semantics is not enough. Existing systems implement
parts of them piece meal; it's the way they are connected up together that is
also important. In particular, we enforce logging at the point of decision in
the exploration library, and we join rewards with delaly in the join service,
and we learn from the data outputted by teh join service. The join service is a
simple abstraction, implementable in a variety of simple ways (we use a
streaming service, one of our customers uses a key-value store, etc.), but it is
an important and powerful abstraction because it whacks away at 2 sources of
technical debt: decoupled data colleciton and a form of bias called reward-delay
bias.
- The result of the above is the Decision Service, which we have built and
deployed, and open-sourced (mostly). To date we have completed 3 production
deployments spanning different companies, \msn.com, Complex, and
\tr. We have also started applying it to health and machine failure
diagnosis. See section X.
- Summarize our results, they've been good. Our eval focuses less on standard
system benchmarks and more on the kinds of technical debt we try to avoid.
- Contribution list
}

\ignore{
Machine Learning has well-known high-value applications, yet effective
deployment is fraught with difficulties in practice~\cite{BPQCCPRSS,hitd}.
Machine-learned models for recommendation, ranking, and spam detection all
become obsolete unless they are regularly infused with new data.  Safely using
data is actually quite tricky--thus the growing demand for data scientists.  How
do we create a machine learning system that safely generates and uses new data?

\begin{figure}[t]
\centering
  \includegraphics[width=.6\columnwidth]{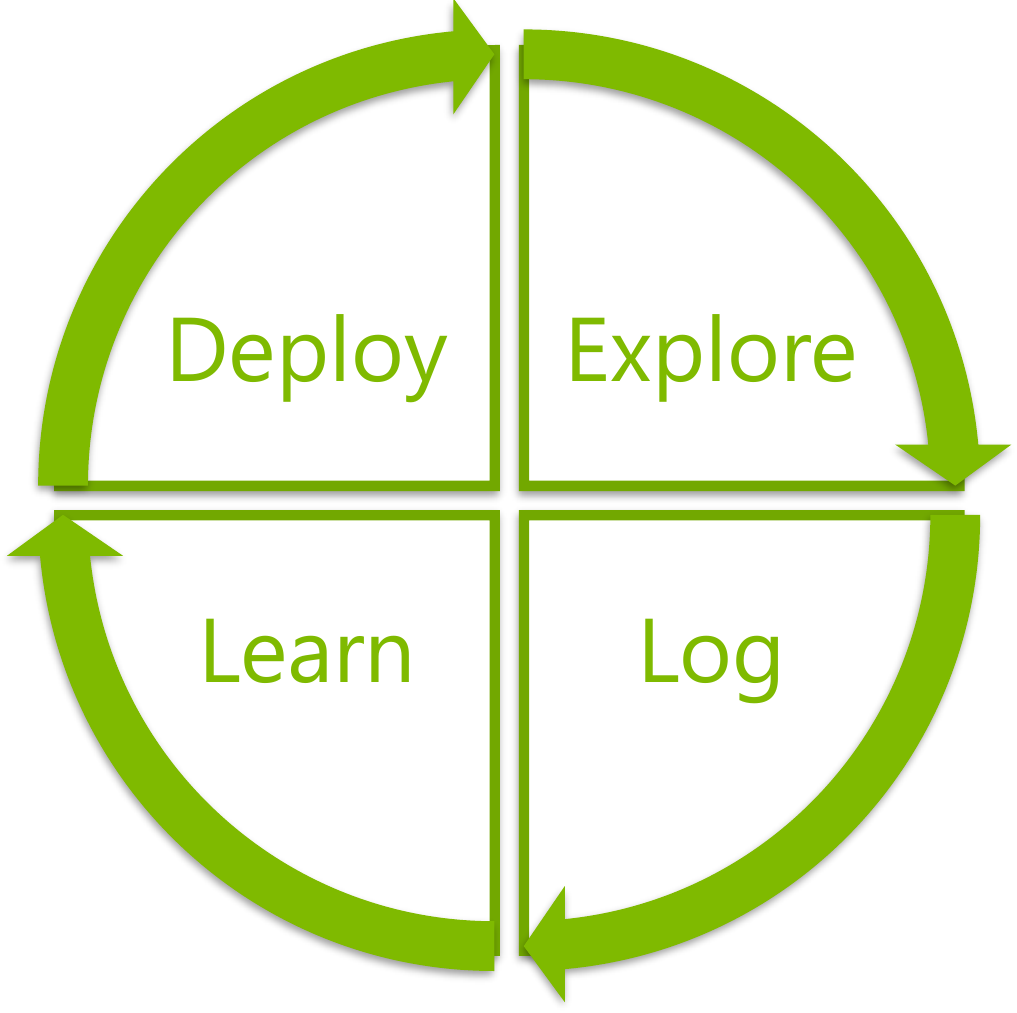}
  \caption{Complete loop for an effective machine learning system.}
  \label{fig:cycle}
\end{figure}

A complete 4-step loop is required: \emph{explore} to collect the data,
\emph{log} this data correctly, \emph{learn} a good model, and \emph{deploy} it
in the application (see \reffig{cycle}). We see two major challenges to
implementing this loop effectively:
\begin{enumerate}
\item Exploration is essential to enable correct learning but is often
  neglected and difficult to get right.
\item Often, each part of the loop is managed independently, resulting
  in the complete process being inefficient, expensive, slow and
  error-prone.
\end{enumerate}
These challenges are subtle and non-trivial to avoid through best
practices alone, yet most machine learning efforts focus on training
and deploying models, neglecting the other two essential steps:
exploration and logging.

Throughout this paper, we discuss a typical application that repeatedly takes \emph{actions} (\eg which type of news article to show) when faced with a particular \emph{context} (\eg user demographics) to obtain a desirable outcome quantified as a \emph{reward} (\eg a click). Machine learning finds a good \emph{policy} mapping contexts to actions, \eg show politics articles to adults, and sports articles to teenagers. As a running example, we use a simple news website (\news) displaying one news article for each visitor.

\xhdr {Exploration and Multiworld Testing.}  Learning can be
biased when the decisions made by an algorithm affect the data
collected to train that algorithm~\cite{BPQCCPRSS,hitd}.  For example,
it is not possible to learn that a news article about sports generates
more clicks than one about politics when shown to a particular
demographic, without showing both types of articles to this
demographic at least some of the time.  Hence, there is a need to
\emph{explore} so as to acquire the right data.  Restated, collecting
a \emph{biased} dataset, no matter how large, does not ensure good
learning~\footnote{Section 2.3 of \cite{BPQCCPRSS} contains a classic
  example of this phenomenon.}.

A common methodology for exploration is
\emph{A/B testing} ~\cite{KohaviAB-2015,KohaviLSH09}: policy A is
tested against policy B by running both live on a small percentage of
user traffic. Outcomes are measured statistically so as to determine
whether B is better than A. This methodology requires data scaling
linearly with the number of policies to be tested and it requires that the
policy to be evaluated is known during data collection.

\emph{Contextual bandits} \cite{monster-icml14,Langford-nips07} is a
line of work in machine learning allowing testing and optimization
over exponentially more policies for a given number of
events\footnote{\Eg in one realistic scenario, one can handle 1
  billion policies for the data collection cost of 21 A/B
  tests~\cite{MWT-WhitePaper-2016}.}.  We refer to this dramatic
improvement in capability as {\bf\em Multiworld Testing} (\MWT). The
essential reason for such an improvement is that each data point can
be used to evaluate all the policies picking the same action for the same context ({\em i.e.}, make the same decision for the same input features
rather than just a single policy as in A/B testing. An important
property of \MWT is that policies being tested do not need to be
approved, implemented in production, and run live for a period of time
 (thus saving much business and engineering effort). Furthermore, the
policies do not even need to be known during data collection.


\OMIT{A typical application repeatedly takes \emph{actions} (\eg which type of news article to show) when faced with a particular \emph{context} (\eg user demographics) so as to obtain a desirable outcome quantified as \emph{reward} (\eg a click). Machine learning can learn a good \emph{policy} mapping from contexts to actions. For example: show politics articles to adults, and sports articles to teenagers. Exploration is there to collect a sufficiently diverse data set to learn from.}

\xhdr{The \MWT \DS.}  Creating a learning system that supports
complete data lifecycle requires solving a number of
technical issues.  How do you record the correct action and context
for learning?  When rewards arrive with a considerable delay after the
actions are chosen, how do you \emph{join} with the corresponding
context and action? And how do you do it with high throughput? How do you
learn from the data and deploy a model quickly enough to exploit the
solution?  And how do you fit all these pieces together into a single
easy-to-use system? Because of these issues, implementing \MWT for a particular
application requires a substantial amount of new programming
\emph{and} considerable infrastructure.  Due to the complexity and
subtlety of creating a correct system\footnote{Although failures are typically
  not published, we know of several failed attempts to employ
  online learning.} tackling this task is well
beyond the capability of most people hoping to apply machine learning
effectively.

Our experience suggests that a dedicated system is the only feasible way to deploy principled experimentation and data analysis.  Many of the pertinent issues are better addressed once and for all, rather than re-developed for each application.  To meet this need, we provide programming and infrastructure support for \MWT via a unified system suitable for many applications. We present a general modular design for the system, incorporating all four components of the learning cycle in Figure~\ref{fig:cycle}. We define the functionality of the components and the ways they interact with one another.  Modularity allows implementation details to be customized to the infrastructure available in a given organization. Using this design, we build our own system for \MWT, called the \emph{MWT \DS}, which includes both the algorithms and the supporting infrastructure. The system consists of several modules which can be used jointly or separately depending on application-specific needs.  The system is fully functional and publicly available.

We deployed the \ds in production to recommend personalized news
articles on a heavily-trafficked major news
site at \msn.  All previous machine learning approaches failed in this setting,
yet the system provided a 25\% relative lift in click-through rate.

In summary, we make the following contributions:
\begin{noindentlist2}
\item We describe a methodology for \mwt that provides strong theoretical
guarantees while addressing many practical concerns (\refsec{mwt}).
\item We present a service for \mwt that eliminates most sources of
error by design. This is achieved through a simple API
(\refsec{api}) and modular component interfaces (\refsec{design}), as well as
techniques for robust randomization and offline reproducibility.
\item We describe our experience deploying the service with \msn in a
  live production environment (\refsec{deployment}), and also evaluate
  performance using systems as well as learning criteria
  (\refsec{eval}).
\end{noindentlist2}

\OMIT{\subsection{Outline of the paper}

In section~\ref{sec:mwt} we discuss the basics of Multiworld Testing and
the machine learning methodology in use.  In section~\ref{sec:api} we
provide examples of the api for application developers.
Section~\ref{sec:design} covers the design goals and architecture of the
system.  Section~\ref{sec:deployment} covers the details of a large scale
deployment at \msn.  Section~\ref{sec:eval} evaluates performance and
ablation of the system.  We close with sections~\ref{sec:related}
and~\ref{sec:future} which discuss related work and future work
respectively.

\alekh{Say something about sections 5 and 6 after they become stable,
  and depending on whether they remain.}}
}

\section{Motivation}
\label{sec:motivation}

Our motivation to create the \ds came from failures and hidden costs both we
and others have encountered\footnote{Unfortunately, such failures are
typically not published.} while deploying contextual learning solutions for
production applications~\cite{Deepak,Langford-www10,Langford-wsdm11,bottou-ad}.
Each application required substantial code and infrastructure, and was subtle
enough that correct implementation eluded most
developers.


Below are the sources of failure that we target, at a high level. If
not addressed by design, most of these issues can cause
difficult-to-debug performance degradation later in the deployment and
lead to technical debts.


\begin{noindentlist2}
  \item[]\noindent {\bf (F1) \asedit{Partial feedback and bias.}} Due
    to partial feedback, a system that does not explore will reinforce
    its own biased decisions. Rewards that arrive after variable
    delays can introduce bias \asedit{in favor of actions with faster
      positive feedback.}  These biases create a discrepancy between
    (predicted) performance at learning time and (actual) performance
    at decision time.  {\bf {\em Examples:}} The baseline systems in
    our content recommendation deployments do not explore, and are
    unable to evaluate actions they do not pick.\footnote{An
      illustration of such problems can be found in the blog post:
      \url{http://hunch.net/?s=randomized+experimentation}}
    \asedit{Fast clicks on ``click-bait" articles can temporarily bias
      the system in their favor.}  \asdelete{Section 2.3
      of \cite{bottou-ad} contains another classic example.}
    \aadelete{ all collect disproportionately more data on content
      they rank highly.}

  \item[]\noindent {\bf (F2) \asedit{Incorrect data collection.}}
    \asdelete{Due to delayed reward feedback, data for contextual
      learning is often collected in a distributed manner. This}
    \asedit{Distributed and delayed data collection} increases
    inconsistencies between the data seen at decision time and the
    data used for learning. With complex pipelines, it is common to
    log not the action chosen by the ML algorithm, but the outcome at
    the end of the pipeline. The benefits of \mwt are lost if the data
    is incorrect.  {\bf {\em Examples:}} \asedit{In \msn and \tr, the
      context features (\eg user browsing history and click
      statistics) can be updated by independent processes that
      pre-date our deployment and are not necessarily designed with ML
      in mind.} \msn editors periodically lock content on the site
    (\eg breaking news) which may override our decisions and typical
    systems record just the editorial override.

\ignore{ All baseline systems in our deployments collected incomplete
  data at the time of decision, \eg to save bandwidth. \msn maintains
  user browsing history via a separate process; \tr maintains click
  counts via a separate process. \msn editors also lock certain
  article slots for breaking news or must-have stories, which would
  alter the effect of a ranking decision.  } \ignore{ Concrete
  examples include: a decision is overriden by a downstream component
  and that decision is recorded instead of the original one; or IDs
  are stored that point to features instead of the features
  themselves, and these features may be updated by separate processes
  over time.  }

  \item[]\noindent {\bf (F3) Changes in the environment.} Real
    environments are {\em non-stationary}: the distribution of inputs
    to the system---\eg user requests or input features generated by
    upstream components---as well as the processing of outputs by
    downstream components, changes over time.  Adapting to these
    changes is necessary to maintain performance.  {\bf {\em
        Examples:}} \asedit{In our deployments,} we have seen breaking
    news events sway \msn users' interests en masse. \asdelete{\tr's
      algorithm for updating click statistics has changed several
      times, in turn changing the input to the content recommendation
      system.}  \asedit{Downstream business rules at \msn and \complex
      \ignore{override `our' content choices and do} also change over
      time.}  \asdelete{\complex has hinted at new downstream business
      rules similar to \msn's that would override content decisions
      (\eg ensuring topic diversity).}

  \item[]\noindent {\bf (F4) Weak monitoring and debugging.}  The
    ability to reproduce an online run offline is key to debugging a
    learning system, but rarely supported in full. \ignore{If the ML
      software is an unknown black box, diagnosing \asedit{subtle}
      bugs in the learning code is extremely difficult.} Most systems
    lack the ability to ask ``what if'', or {\em counterfactul}
    questions, making it hard to implement safeguards or automated
    responses even when real-time monitoring is available.  {\bf {\em
        Examples:}} Early in our \msn deployment, the cause for a
    buggy ML model was correctly traced only after full
    reproducability was achieved.  All product teams we engaged with
    used expensive A/B tests to monitor the performance of alternative
    solutions.

\ignore{  \item[]\noindent {\bf (F5) \asedit{Culture/process}.} The process of
    creating an ML system can also be the source of \asedit{issues
      and} technical debt. For example, a disconnect between system
    developers and ML researchers providing software can lead to
    excessive glue code around the (black-box) ML software. When an ML
    system replaces a function performed by humans, this can create
    \asedit{unwanted feedback loops which}
  make it difficult to isolate gains provided by
  the system.
  {\bf {\em Example:}} \msn editors, who provide the baseline ranking of
  content, have increasingly been using the \ds's
  ranking to \asedit{improve} their own.}
\end{noindentlist2}


\begin{table}[!t]
{\scriptsize
\begin{center}
\begin{tabular}{|l|l|}
\hline
{\bf Category} & {\bf Example ML systems} \\
\hline
Supervised & Caffe, CNTK, GraphLab, LUIS, Minerva, mldb.ai, MXNet
\\ learning & MLlib, NEXT, Param.Server, ScikitLearn, TensorFlow, Torch \\
\hline
Bandit learning & Clipper, Google Analytics, LASER, VW, Yelp MOE
\\
\hline
Cloud service & Amazon ML, AzureML, Google Cloud ML \\
\hline
A/B testing & Google Analytics, MixPanel, Optimizely \\
\hline
\end{tabular}
\end{center}
\caption{A simplified categorization of ML systems (overlaps exist,
\eg cloud services support supervised learning). None of the systems address our main
technical debts (\refsec{motivation}).}
\label{tbl:mlsystems}
}
\end{table}

There is a common theme above of ensuring consistency between
learning-time and decision-time performance. This is not surprising:
being able to accurately predict online performance is
\asedit{essential to} a learning system.

\ignore{
Although failures like the above are not typically published, we have
experienced all of the above technical debts in real deployments in the
past, and know of others experienced by our colleagues.  T1 has plagued a
variety of online applications we have worked on (\eg \reftbl{apps}), and is the
primary reason we are contacted for help. Section 2.3 of \cite{BPQCCPRSS}
contains a classic example of this phenomenon. T2 arises in countless creative
ways, ranging from programmers incorrectly logging exploration information as
features, to features being changed between decision time and learning time, to
failure in upstream components affecting what is seen during decision time and
learning time. T3 manifests itself in a system not being able to maintain
the improvements it initially provided, or otherwise behaving differently
than in the past, and is often only visible using slow and expensive A/B
testing. T4 has cost countless hours debugging unexpected system and learning
algorithm behavior from inadequate data; also, the inability to ask powerful
questions, such as how well would the baseline policy have done during this time
period, makes it difficult to get timely answers that could be used for
real-time response or safeguarding. T5 often manifests in a competitive culture
around who is providing what gains to the system; sometimes competition is good,
but in this case it can hurt the statistical significance of an ML system's
gains.
}

\asedit{We discuss related work in \refsec{related}, but note here
  that the existing ML systems fail to address most of these
  \asedit{issues}.  In particular, the supervised learning systems
  listed in \reftbl{mlsystems} do not support exploration, and hence
  do not address contextual learning settings or the issues in (F1).}
Several systems support bandit learning but not {\em contextual}
bandit learning, which is significantly more powerful.  \asdelete{(\eg
  we had to modify VW to support contextual bandit exploration).}
None of the systems support data collection to properly address (F2),
except LUIS (for a different setting of active learning) and NEXT,
which additionally handles multi-armed bandits but not general
contextual bandits and neither system supports offline
evaluation/monitoring highlighted in (F4). The cloud services support
retraining and deploying models, but not in an online
fashion. Clipper~\cite{clipper} incorporates feedback in real-time and
thus better addresses (F3), but only explores over the predictions of
existing (batch-trained) models. Several systems provide \asedit{some}
debugging functionality, and some (\eg NEXT) support reproducible
runs, addressing (F4). \asedit{While A/B testing platforms can answer
  counterfactual questions by running a live experiment for each
  question, the systems in \reftbl{mlsystems} cannot answer new
  counterfactual questions from already-collected data, let alone
  match the sample efficiency of \mwt.}  The \ds fills all these gaps.

The paper addresses the motivating issues as follows.  The ML
  methodology in \refsec{mwt} partially addresses (F1), (F3) and (F4),
  and the system abstractions in \refsec{design} aim to address all
  issues at once. Our evaluation (\refsec{techdebt-eval}) investigates
  the specific examples listed above (among others). \ignore{While our system
  does not directly address (F5), we discuss how we handled
  process-related issues in \refsec{future}.}

\ignore{
- Although our focus and contributions are ML and sys to reduce these debts, we
need to meet our customer demands for latency and scalability too. But we don't
focus too much on this in the paper.
}

\ignore{
- motivation came from prior failures to deploy such a service and the
tecnnical debts we observed over time. We list these debts below; this list will
serve as the guideline for our ml synthesis and system design, and we will often
refer back to these numbers when addressing the debts. Without further ado\ldots
\\
- We or our colleageus have experienced real failures in each of these
categories. For example, \ldots \\
- In addition, existing systems fail to address these technical debts. We have
summarized existing systems according to their ability to support \\
contextual learning and their ability to address these debts. [How not to offend
anyone?] \\
- Although our focus and contributions are ML and sys to reduce these debts, we
need to meet our customer demands for latency and scalability too. But we don't
focus too much on this in the paper.

- Technical debt (table summarizing which ones solved + name of technique and
 forward ref? These could subsume the design requirements, so number them and
 refer to them in Design section; to that end, could title each one with the
 solution/need rather than the problem. In each, summarize examples of how
 things can go wrong.

- change in environment requires slow A/B testing, can only ask weak questions
and compare to past performance, not current (also not reproducible), downstream
teams and lack of communication etc..

- Then say that we address them via a combination of a synthesized ML methodology and our system design, discussed in next two sections.

To mention:

subtlety of creating a correct system\footnote{Although failures are typically
  not published, we know of several failed attempts to employ
  online learning.} tackling this task is well
beyond the capability of most people hoping to apply machine learning
effectively.

\footnote{Section 2.3 of \cite{BPQCCPRSS} contains a classic example of this
phenomenon.}.
} 

\section{Machine learning methodology}
\label{sec:mwt}

\begin{table*}[th]
{\footnotesize
\begin{center}
\begin{tabular}{l|l|l|l}
& News website (\news) & Tech support assistance (\techsupport) & Cloud controller
(\job)
\\
\hline {\bf Decision to optimize} & article to display on top & response to query & wait time before reboot unresponsive machine \\
\hline
{\bf Context} & location, browsing history,... & previous dialog elements & machine hardware/OS, failure history,...
\\
{\bf Feasible actions} & available news articles & pointers to potential solutions & minutes in
\{1,2,...,9\}
\\
{\bf Reward} & click/no-click & (negative) human intervention request & (negative)
total downtime
\\
\hline
\end{tabular}
\end{center}
\vspace{-5pt}
\caption{Example applications of the \ds. Each is representative
of a real deployment discussed in \refsec{deployment}.}
\label{tbl:apps}
}
\end{table*}

From the machine learning perspective, we implement a capability we
call \emph{multiworld testing} (\MWT), which can test and optimize
over $K$ policies using data and computation that scales as $\log K$,
without any prior knowledge of the policies.  We
show that \mwt addresses (F1) and helps address (F4).

Our methodology synthesizes ideas from contextual bandits (\eg
\cite{monster-icml14,Langford-nips07}) and policy evaluation (\eg
\cite{Langford-www10,Langford-wsdm11}).  The
methodology is modular: exploration and logging support offline
evaluation and learning over arbitrary policy sets. This modularity
maps to modularity in our system design.



\ignore{
\mwt is based on a line of theoretical research on contextual bandit
learning~\cite{Langford-nips07,Langford-wsdm11}. The basic framework and
techniques draw from this literature with the goal of creating a
methodology that is suitable for a real, general-purpose system. We overview
the methodology here.
}


\ignore{
- Define the theoretical model and summarize typical guarantees \\
- Explain why it is the right model for real-world decisions made by
systems/services \\
}

\xhdr{Contextual decisions.}
Consider an application \APP that interacts with its environment, such as a news
website with users or a cloud controller with machines. Each interaction
follows the same broadly applicable protocol:
\vspace{-1mm}
\begin{enumerate}
\item A \emph{context} $x$ arrives and is observed by \APP.
\item \APP chooses \emph{action} $a \in A$ to take ($A$ may depend on $x$).
\item A {\em reward} $r$ for $a$ is observed by \APP.
\end{enumerate}
\vspace{-1mm} \reftbl{apps} shows examples from our deployments.
Contexts and actions are usually
represented as feature vectors.  \APP chooses actions by applying a
\emph{policy} $\pi$ that takes a context as input and returns an
action. The goal is to find a policy that maximizes average reward over a
sequence of interactions.
We assume for now that \APP faces a stationary environment.\footnote{Stationary
here means the context and the reward given the context-action pair are drawn independently from fixed distributions.}

\xhdr{Exploration and logging.}  An \emph{exploration policy} is used
to randomize each choice of action. The randomization need not be
uniform and for best performance should not
be~\cite{monster-icml14,bandits-exp3}.  A simple policy is \EpsGreedy:
with probability $\eps_0$ it chooses an action uniformly, and uses a
\emph{default policy} $\pi_0$ otherwise. $\pi_0$ might be the baseline
deployed in production or the current best guess for an optimal
policy. $\eps_0$ controls an \emph{explore-exploit tradeoff}: $\pi_0$
guarantees a minimum performance while randomization explores for
better alternatives.  The parameters $\eps_0$ and $\pi_0$ can be
changed over time.  The contextual bandit literature provides several
exploration policies including near-optimal schemes.

Each interaction is logged as a tuple $(x,a,r,p)$, where $p$ is the
exploration policy's probability of choosing $a$ given $x$. These datapoints are
called \emph{exploration data}.  Recording $p$ enables unbiased policy learning, which addresses (F1).

\xhdr{Policy learning.}
Given $N$ exploration datapoints, we can \emph{evaluate} any policy
$\pi$ (\ie estimate its average reward) regardless of how the data was
collected. The simplest approach is to use \emph{inverse propensity scoring}
(\ips):
\vspace{-2pt}
\begin{align}
\label{eq:ips}
\muIPS(\pi)
= \textstyle \frac{1}{N} \sum_{t=1}^N
	\; \indicator{\pi(x_t) = a_t} \; r_t / p_t,
\end{align}
where the indicator is $1$ when $\pi$'s action matches the exploration
data and $0$ otherwise.  This estimator has three important
properties. First, it is \emph{data-efficient}: each interaction can
be used to evaluate any $\pi$ that has a matching action,
\emph{regardless of the policy collecting the data}.  In contrast, A/B
testing only uses data collected using $\pi$ to evaluate
$\pi$. Second, the importance weighting by $p_t$ makes it
statistically \emph{unbiased}: it converges to the true reward as
$N\to\infty$.  Third, the estimator can be \emph{computed
  incrementally} as new data arrives.

Thus, using a fixed exploration dataset, we can compute accurate
counterfactual estimates of how arbitrary policies \emph{would have
  performed} without actually running them, in real-time.  We use this
in our system design to enable advanced monitoring and safeguards,
addressing (F4). In contrast, A/B testing would have to run a live
experiment to test each policy.

The ability to reuse data is what makes this approach exponentially
more efficient than A/B testing, in a manner we can quantify. Suppose
we wish to evaluate $K$ different policies. Let $\eps$ be the minimum
probability given to each action in the exploration data (for
\EpsGreedy, $\eps =\eps_0/|A|$), and assume all rewards lie in
$[0,1]$. Then, with probability $1-\delta$ the \ips estimator yields a
confidence interval of size $\sqrt{\tfrac{C}{\eps
    N}\log\tfrac{K}{\delta}}$ for the values of all $K$ policies
simultaneously, where $C$ is a small constant. Crucially, the error
scales $O(\log K)$. In contrast, with A/B testing the error could be
as large as $C\sqrt{\tfrac{K}{N} \log\tfrac{K}{\delta}}$, which is
{\em exponentially worse}. This also underscores the need to explore
over all relevant actions: if $\epsilon = 0$, we cannot correctly
evaluate arbitrary policies.

Policy evaluation allows us to search a policy class $\Pi$ to find the
best policy, with accuracy similar to the above bounds (replace $K$
with $|\Pi|$). This is called {\em policy training}. Typically $\Pi$
is defined by a tunable template, such as linear vectors, decision
trees, or neural nets.  Note that we need not test every policy in
$\Pi$; instead, we can use a reduction to \emph{\CSC}~\cite{DR11}, for
which many practical algorithms exist.

\ignore{

Let us compare the statistical efficiency of \mwt to that of A/B testing.
Suppose $N$ data points are collected using an exploration policy which places
probability at least $\eps$ on each action (for \EpsGreedy, $\eps
=\eps_0/\text{\#actions}$), and we wish to evaluate $K$ different policies. Then
the \ips estimators for all $K$ policies have confidence intervals whose width
is $\sqrt{\tfrac{C}{\eps\, N} \log\tfrac{K}{\delta}}$,  with probability at
least $1 - \delta$, where $C$ is a small absolute constant.%
\footnote{This holds for any $\delta>0$ and any $N>\tfrac{1}{\eps}$,
  by an application of a \emph{Bernstein's Inequality}, a well-known
  tool from statistics.}  This is an {\bf\em exponential} (in $K$)
improvement over A/B testing since an A/B test of $K$ policies
collecting $N$ data points has confidence intervals of width
$C\sqrt{\tfrac{K}{N} \log\tfrac{K}{\delta}}$. This also shows the necessity of exploration for policy
learning. If $\epsilon = 0$, we cannot correctly evaluate arbitrary
policies.

\emph{Policy training} finds a policy that approximately maximizes the
estimated reward (usually given by the \ips estimator). A naive but
computationally inefficient solution is to evaluate every allowed
policy and output the one with highest estimated reward. A better
approach is to use a reduction to \emph{\CSC}~\cite{DR11}, for which
many practical algorithms have been designed and implemented. The
choice of algorithm implicitly defines the class of allowed policies:
for example, there are algorithms for policy classes specified by
linear representations, decision trees, and neural
networks.\footnote{Even though the prior discussion considers only
  finite number $K$ of policies, conclusions extend to continuous sets
  such as linear representations using standard statistical arguments
  (see e.g.~\cite{Langford-nips07}).} All such policies tend to be
very fast to execute.  Regardless of the policy training procedure,
the confidence bound above implies that the expected reward of the
trained policy tends to its estimated reward using \ips.
}

\xhdr{Non-stationarity.}  The estimates of policy performance are only
predictive of future performance if the environment is stationary, but
in practice applications exhibit only periods of (near-)stationarity.
To cope with a changing environment, our system design implements a
continuous loop in the vein of \reffig{cycle}. This has two
implications for policy training.  First, an online learning algorithm
is (almost) necessary, so that new datapoints can be incorporated
quickly without restarting the training. Second, since most online
learning algorithms gradually become less sensitive to new data (via a
shrinking learning rate in the optimization), we periodically reset
the learning rate (\eg for \msn we reset each day) or use a constant
rate throughout (\eg for \complex), thereby partially addressing (F3).


\ignore{
We use the informal term \emph{stationarity timescale}: the time interval during which expected
rewards do not change much. To cope with a changing environment,
we use a continuous loop in the vein of \reffig{cycle}. This has two
implications for policy training. First, an online learning algorithm
is (almost) necessary, so that new datapoints can be incorporated
efficiently without restarting the training. Second, while most online
learning algorithms gradually become less sensitive to new data, we
set the sensitivity according to the stationarity timescale. This
means that learning stabilizes during the timescale, but can adapt to
new trends over longer periods of time.  This can be achieved by
occasionally resetting the step-sizes present in most online learning
algorithms.
}



\ignore{
\sid{cut some of this to save space}
Within the timescale, we need enough completed
interactions to enable policy learning. A rare but crucial reward
outcome---\eg clicks are rare---is a learning bottleneck. A good
rule of thumb for learning (with a linear representation) is that the
number of rare outcomes in the timescale should be much larger than
$\algofont{\#actions} \times \algofont{\#features}$.

Non-stationarity can often be partially mitigated by choosing features
with a more persistent effect, and/or a learning algorithm that adapts
in a changing environment.
}

\xhdr{Problem framing.}
``Framing'' the problem---defining the context/features, action set, and reward
metric---is often non-trivial and is a common difficulty in many applications of
machine learning. The \ds does not directly address problem framing, but eases
the task in two ways: auto-generating features for content recommendation
applications (\refsec{auto-features}), and allowing testing of different
framings (including changes to the reward metric, with proper logging)
without collecting new data (see ``Offline Learner", \refsec{arch}).


\ignore{
In practice, there may be several reasonable ways to define rewards, especially
when the outcome of an interaction consists of multiple observations such as a
click and dwell time.  Additionally, the reward metric often is changed or
refined over time, and auxiliary metrics may be added.  Our methodology does not
require committing to a particular reward metric during exploration; switching
to a new reward metric is only a simple post-processing of the exploration
dataset as long as the required observations are logged.
}

\ignore{
\sid{Cut but briefly mention in msn deployment discussion}
In some applications, a set of ordered actions (a \emph{slate}) such as a ranked
list of search results or news articles is chosen in each interaction. Treating
the entire slate as a single action is inefficient, because too many slates are
possible. One practical approach, adopted in our \msn deployment, is to
implement \mwt for the top slot, and use the learned policy for other slots. A
more principled approach explores and learns for all slots at once
\cite{AkshayCB-2015}.
}

\section{System Design}
\label{sec:design}

\asedit{While our ML methodology breaks the biased feedback loop in
  (F1) and enables arbitrary policy evaluation, it takes systems
  support to fully address changing environment, data collection
  issues, and delay-related bias.} For example, the ML methodology
simply assumes $(x,a,r,p)$ data is correctly provided as input, but
this is nontrivial in practice.

Our system design fills these gaps to operationalize our ML
methodology.  We list our design goals (\refsec{goals}), define a set
of abstractions and an architecture that implements them
(\refsec{arch}), and highlight key techniques to meet our goals
(\refsec{techniques}).

\subsection{Design goals}
\label{sec:goals}

Our goals are addressing the twin concerns of keeping a low
technical debt and having a performant system.

\ignore{
The \ds is designed to implement a complete machine learning loop and
harness the power of \mwt for a wide range of potential applications,
while preventing many of the errors we have encountered deploying in
practice. Below we describe the associated challenges and design
requirements.
}

\xhdr{\asedit{Failures / technical debt.}} So far we have discussed
these issues at a high-level. We now create concrete design goals
motivated by specific examples.

\emph{Logging at the point of decision:} \MWT relies crucially on
accurate logging of the $(x,a,r,p)$ tuples. The system must record
$(x,a,p)$ at the time of decision and match the appropriate reward $r$
to it. There are many ways we have seen this go wrong in practice. For
example, features may be stored as references to database entries that
are updated by a separate process. Consequently, the feature values
available at learning time might differ from those at decision time,
whether due to updates/additions/removals or access failures. When
optimizing an intermediate decision in a complex system, the action
chosen initially might be overridden by downstream business logic
(like editorial locking), and this is the action that gets logged. In
this case, the probabilities $p$ correspond to the choice of $a$, not
the recorded action, and are thus incorrect. Sometimes, probabilities
are stashed as part of the context and included as an
action feature by accident.

\emph{Experimental unit for joining:} Different rate of reward
arrivals on different actions can lead to biases as discussed in
\refsec{motivation}. This can be avoided by waiting a pre-set duration
for reward arrival before the joining is done, and assigning a default
reward (say $0$) if no feedback arrives in that duration.

\emph{Continuous learning:} A system that does not continuously learn
(or fails to reset the learning rate as specified by our methodology)
does not adapt well in non-stationary environments where a
low-latency learning loop is required.

\emph{Reproducibility:} Interactive systems which span many components
are challenging to debug. Events may be delayed, reordered, or dropped
and affect the system in complex ways, making it difficult to
reproduce a bug. The difficulty is magnified when the system is
continuously learning, because the system is no longer stationary, and
it is difficult to disentangle issues in the learning algorithms from
systems issues. The ability to fully reproduce an online run offline
is essential to effectively debug learning.

\ignore{
x no user profile to some
x features added/removed failures
x reward delay bias some actions appear before others, in a continuous learning
x editors override downstream.
x changes in external world
x full reproducibility for debugging
- safeguard triggers
}

\ignore{
\xhdr{Data collection.}  \MWT relies crucially on accurate logging of
the $(x,a,r,p)$ tuples. The system must record $(x,a,p)$ at the time
of decision and match the appropriate reward $r$ to it. There are many
ways we have seen data collection go wrong in practice. For example,
the probabilities $p$ may be incorrectly recorded or accidentally
included as action features.  Features may be stored as a reference to
a database which is updated. Consequently, the feature values
available at the time of policy evaluation might differ from the ones
at the time the interaction was recorded. When optimizing an
intermediate step in a complex system, the action chosen initially
might be overridden by downstream business logic, and the recorded
action might incorrectly be this final outcome rather than the
initially chosen action. Finally, the rewards, which are often delayed
and may arrive from entirely different paths within a given
application, may be lost or incorrectly joined to the decision.

\xhdr{Machine learning.}  The service should provide adequate systems
support for machine learning, which means supporting a variety of
algorithms and processes used by data scientists in practice to
explore, tune, and adapt policies. As much as possible, we would like to
not be tied to a particular learning library or policy structure, so
that applications can add the Decision Service on to their existing ML
workflow.  Also, in order to adapt to a changing environment and
workloads the \ds should be able to revise the learned policies
continuously and support real-time resets/reconfigurations of the learning
algorithms. Furthermore, there is plenty of prior art in machine
learning (\eg supervised learning) that should be leveraged when appropriate. Finally, the service should support \emph{offline
  experimentation} (using exploration data instead of additional live
experiments) to tune and try out exploration/learning algorithms in
realistic conditions.
}

\xhdr{Systems goals.} In addition to supporting the technical debt
minimization goals above, the system needs to meet the performance and
functionality demands of our customers.

For interactive high-value applications, serving latency tends to be directly
linked to user experience and revenue~\cite{Schurman2009}.  To optimize these
reward metrics, the system must provide decisions in \about{10ms} or less to
keep application response times under \about{100ms}. Some applications need to
quickly incorporate new data into the machine-learned policy (\eg for breaking
news stories), so the system needs a learning loop that can update the policy
every few minutes.

As with any service, the system should be scalable and
fault-tolerant. Two issues are unique in our setting. First, delayed
rewards increase the active state of the system where the volume of
state depends on the size of the outcome observation, the interaction
arrival rate, and the typical delay until the reward is
observed. Second, ensuring reproducibility requires care when handling
reordered events from scale-out components or failures. The system
should not lose data that was trained on and should recover the
previously learned policies and other valuable state of the learning
algorithm.

The system should provide flexible deployment options and be easy to use and
customize.  A modular design with well-defined interfaces admits multiple
implementations, allowing us to adapt and evolve the system to
applications' needs. The system should expose the power of our ML methodology
to support offline experimentation, \ie using exploration data (instead of live
experiments) to tune parameters, try other exploration/learning
algorithms, etc.. To reduce setup
complexity, sensible defaults should be provided for all components.

\ignore{
\xhdr{Latency.} For interactive high-value applications serving
latency tends to be directly linked to user experience and
revenue~\cite{Schurman2009}.  Many applications now seek to keep their
total response times within 100 ms, so the \ds should make decisions
in 10 ms or less.

Some applications must quickly incorporate new data into the
machine-learned policy (e.g. for breaking news stories) so the \DS
should be able to update the policy every few minutes.

\xhdr{Scalability.} The \ds needs to efficiently support
  applications with both high and low volumes of data. The volume of
  in-memory data depends on the size of the interaction tuple, the
  interaction arrival rate, and the typical delay until the reward is
  observed. The online learning algorithm  should be able to handle the data
  arrival rate. Finally, reproducibility may demand additional
  buffering due to reordering from scale-out components.


\xhdr{Usability and flexibility.}  The \DS should be \emph{modular} so
as to integrate easily with an application's existing
infrastructure. That is, it should consist of components which are
usable independently or in combination with well-defined interfaces
that admit multiple consistent implementations. This avoids costly
re-implementation of existing functionality, and allows the service to
improve seamlessly as better implementations become
available. Supporting multiple programming languages and avoiding
mandatory dependencies on particular external libraries reduces the
barrier to adoption, as applications use different programming
languages and environments, and some hesitate to take additional
dependencies.  Finally, the service should be easy to try out, since
application developers may be wary of automatic learning
systems. Overall, the design should not enforce a ones-size-fits-all
approach, but provide sensible defaults for every component to reduce
setup complexity for common cases.

Finally, the service should support \emph{offline
  experimentation} (using exploration data instead of additional live
experiments) to tune and try out exploration/learning algorithms in
realistic conditions.

\xhdr{Robustness.} The application must continue making decisions
despite any failures in the components of the \DS. Conversely, the \DS
must be able to recover from the failures, restarts or
reconfigurations in the application itself. Even in the
business-as-usual mode, data collection may be skewed in time, and
some of the data may not reach the service. Either way, the \DS should
not lose data that was received, and should recover the previously
learned policies and other valuable state of the learning algorithm.
}

\subsection{Abstractions and architecture}
\label{sec:arch}

\begin{figure}
\centering
\includegraphics[scale=0.176]{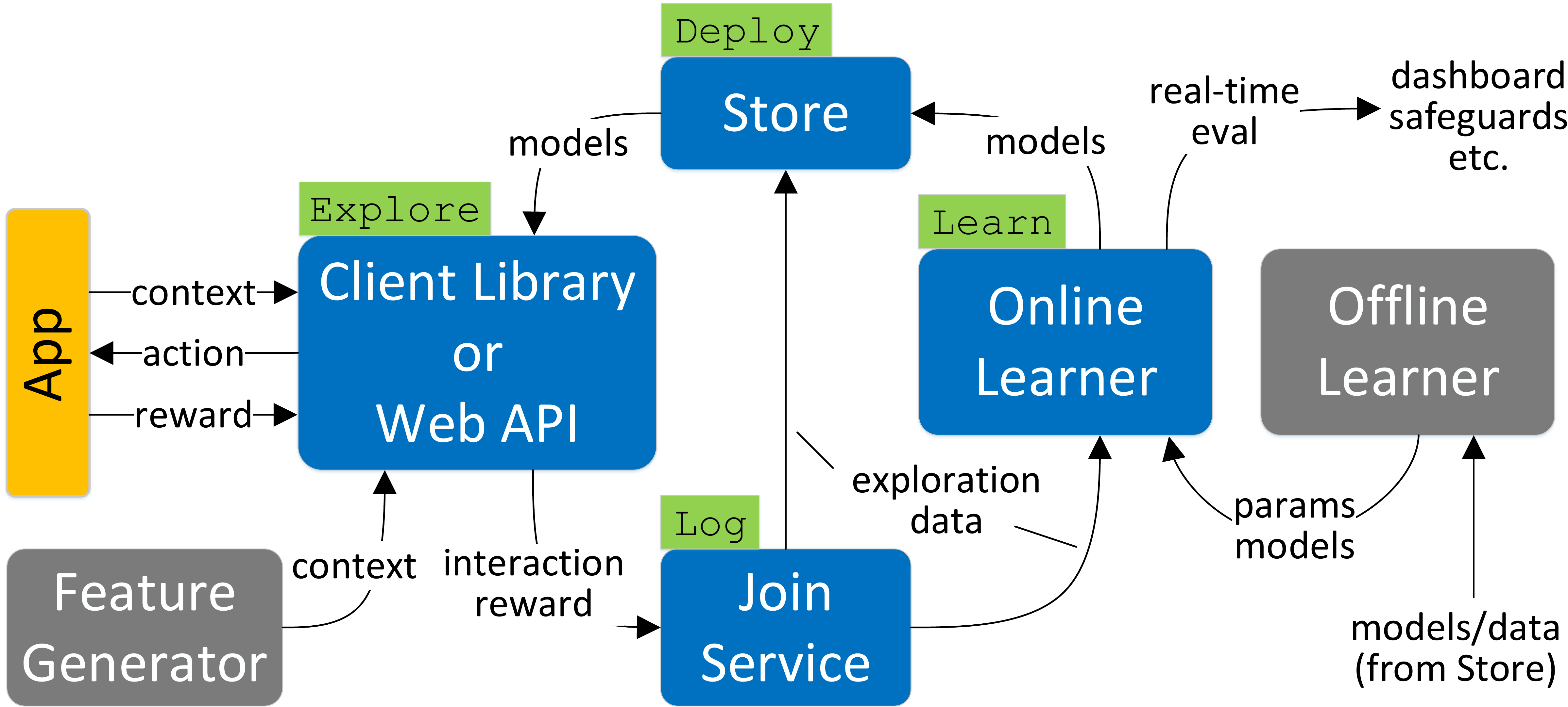}
\caption{\ds architecture}
\label{fig:arch}
\end{figure}

Our system is designed to match the modularity of the ML methodology
including the exploration and logging components and the policy
evaluation and training steps. Doing so, we define an abstraction for
each step of the loop in \reffig{cycle}:

\begin{noindentlist2}
  \item \algofont{Explore}: This component interfaces with the \APP.
    It takes as input context features $x$ and an event key $k$ from
    \APP, and outputs an action $a$.  Separately, a keyed tuple
    $\langle k, (x,a,p) \rangle$ is sent to the \algofont{Log}
    component, where $p$ is the probability of the chosen action
    according to the exploration policy. Later, a reward $r$ and key
    $k$ are input from \APP, triggering a transmission of
    $\langle k, (r) \rangle$ to the \algofont{Log} component.

  \item \algofont{Log}: This component generates exploration data by joining
  each $(x,a,p)$ tuple with its reward $r$.  It defines a
  configurable parameter, \emph{experimental unit}, that
  specifies how long to wait for a reward to arrive.  Thus, the component
  takes a stream of keyed observations $(k,o)^*$ and emits a stream of joined
  observations $(o_1,o_2,...)^*$, where observations are joined if they share
  the same key $k$ and appear within time $[t, t+exp\_unit]$, where $t$ is the
  time $k$ was first observed.

  \item \algofont{Learn}: This component performs policy training \emph{online}.
  It takes a stream $(x,a,r,p)^*$ of exploration datapoints from the
  \algofont{Log} component and outputs a unique model $(id, model)$. A new model
  can be output after each datapoint.

  \item \algofont{Deploy}: This component stores models via a get/put
    interface.  The \algofont{Learn} component puts $(id, model)$ into
    the store and the \algofont{Explore} component gets a $model$
    using its $id$ (or the most recent model if no $id$ is specified).
\end{noindentlist2}

The abstractions by themselves address several of our
technical debt reduction goals.
\reffig{arch} shows the architecture of the \DS and the
component that implements each abstraction. Each component is a scalable
service.

The {\em Client Library} implements the \algofont{Explore}
abstraction. It implements various exploration policies from the
contextual bandit literature, addressing (F1), and logs the correct
data to the \algofont{Log} component, addressing (F2). In particular,
this enforces logging at the decision point (\refsec{goals}). The
Client Library downloads models from the \algofont{Deploy} component
at a configurable rate; before the first model is available, a
user-provided default policy/action may be used. The Client Library
can be linked to directly for low latency decisions, or accessed via a
portable, scalable web API (see \reffig{api}).

The {\em Join Service} implements the \algofont{Log} abstraction by
joining rewards to decisions, producing correct exploration data to
address (F2).  It also enforces a uniform delay ($exp\_unit$) before
releasing data to the \algofont{Learn} component to avoid
delay-related biases, addressing (F1) (as discussed in \refsec{goals}
above).  The exploration data is also copied to the Store for offline
experimentation (discussed below).

The {\em Online Learner} implements the \algofont{Learn} abstraction
using any ML software that provides online learning from contextual
bandit exploration data. It incorporates data continuously and
checkpoints models to the Deploy component at a configurable rate,
addressing our goal of continuous learning and handling (F3). The
Client Library must use compatible ML software to invoke these models
for prediction.  The Online Learner uses \refeq{ips} to evaluate
arbitrary policies in real-time, realizing \mwt capability. These
statistics enable advanced monitoring and safeguards
(\refsec{safeguards}), addressing (F4).

\alekh{Rename to ``Offline Analytics''?}

The {\em Store} implements the \algofont{Deploy} abstraction to
provide model and data storage. This data is also used by the {\em
  Offline Learner} for offline experimentation, such as tuning
hyperparameters, evaluating other learning algorithms or policy
classes, changing the reward metric, etc.  Our ML methodology
guarantees these experiments are counterfactually
accurate~\cite{Dudik-uai12,DR11,Langford-www10,Langford-wsdm11}.
Improvements from offline experimentation can be integrated into the
online loop by simply restarting the Online Learner with the new
policy or settings. This addresses some usability and
customizability goals.

The {\em Feature Generator} eases usability further by auto-generating
features for certain types of content (\refsec{auto-features}).

\ignore{
\ignore{
The Client Library interfaces with the application, makes decisions, performs
exploration and issues logging requests. The Join Service is
responsible for joining and logging exploration data. The
Online/Offline Learner components perform policy learning, and
optimized policies are deployed back to the Client Library.
}

\ignore{
Each module below has a well-defined interface, meaning each component
can be used in isolation or replaced with customized implementations
to suit the application's environment. For example: the Client Library supports custom logging classes, which can
send interaction data to an
external learning system; the Join Service can be implemented by any
key-value store that supports expiration timers; the Online Learner
can take $(x,a,r,p)$ tuples generated by an external source, and can
be implemented using any ML package that understands this data. We
leverage this flexibility in our deployment with \msn.
}

\sid{Argue client library need not depend on anyone to collect the
right context data, since it logs exactly what was seen by scorer.}
\xhdr{Client Library:} This module correctly implements various
exploration policies from the contextual bandit
literature. Concretely, it takes as input context features $x$ and an
event key $k$ from \APP, and outputs an action $a$.  Separately, a
keyed tuple $\langle k, (x,a,p) \rangle$ is transmitted to the Join
Service, where $p$ is the probability of the chosen action $a$
according to the exploration policy. Later, a reward $r$ and
key $k$ are input from \APP, triggering a separate transmission of
$\langle k, (r) \rangle$ to the Join Service. The exploration
  policy is defined based on the Client Library configuration and can
  depend on a default action, a default policy and/or the output of
  the Online Learner.

\sid{Our design systematically eliminates such bugs by ensuring the $(x,a,p)$
tuple is set aside for logging at the point of decision and is
correctly joined with the reward later. This guarantees the feature
values used for policy evaluation are consistent with the interaction,
the correct probabilities are recorded and the action chosen by the
randomization is correctly logged even if downstream logic overrides
it.}

\ignore{We have implemented versions of the client library in C\#, C++, and Java. The C\#
library is about 5K lines of code: 1.5K for the various exploration policies and another
1.5K which handle batched uploads to the Join Service. The Client
Library can optionally link to Vowpal Wabbit (\vw) \cite{VW}, an open-source online learning library, to invoke policies. The Client Library code is open source~\cite{ds-clientlib}.}

\xhdr{Join Service:} This logging module collects exploration
data: each $(x,a,p)$ tuple is joined with the reward $r$ from the same
interaction, often arriving after a substantial delay or from an
entirely different component. How long to wait for the reward to
arrive (\emph{experimental duration}) can be configured to suit a
particular application. The Join Service thus takes a stream of keyed
observations $(k,o)^*$ and emits a stream of joined observations
$(o_1,o_2,...)^*$ where observations are joined if they share the same
key and occur within this experimental duration. In particular, when a
key is first observed a timer is started.  All observations with the
same key are joined and emitted when the timer reaches zero.  The
joined tuples $(x,a,r,p)$ are then output to the Online Learner (via a
queue), and also sent to storage for offline learning.

 \ignore{We have implemented the Join Service using Azure Streaming Analytics (ASA),
which allows us to specify a delayed join with 4 lines of query
language---the delay is set to the experimental duration.  ASA can be
scaled up/down and handles failures/restarts by storing incoming data
in a fault-tolerant queue (an Azure Event Hub), and replaying data as
needed.}

\xhdr{Online Learner:} This component performs policy training
\emph{online}, \ie quickly updating the policy to incorporate a stream
$(x,a,r,p)^*$ of examples output by the Join Service. It can run
continuously and adapt to changing workloads or application
environments.   The Online Learner also evaluates arbitrary policies in
real time, including the current trained policy, using
\refeq{ips} on the exploration data.  We use these statistics for
safeguards (\refsec{safeguards}) and to display performance on a
dashboard~\cite{PowerBI}.

The Online Learner, in a bit more detail, consists of a Reader Module
to process the data stream, and an ML Module for policy
training. In principle, any ML package providing online
learning from contextual bandit exploration data can be used for the
ML Module.  Policies are generated and checkpointed at a configurable
rate and pulled by the Client Library at another
configurable rate.

\ignore{The Online Learner is implemented in 1.8K lines of C\# as a
stand-alone worker role. It consists of a Reader Module and ML Module.
The Reader is specific to the ML Module: it
translates the interaction data from the on-the-wire JSON format used
by the Client Library and Join Service to a format the ML package can
understand.

In principle, any ML package that provides online
learning from contextual bandit exploration data can be used for the
ML Module. We currently use \vw, which
implements this via reduction to cost-sensitive classification, as per
our ML methodology.  Policies are generated and checkpointed at a configurable
rate and pulled by the Client Library at another
configurable rate.

 By default, we train a policy with linear
representations (a vector of weights) for the latter problem although
many other representations are available.  \vw also supports parallel
learning using the \algofont{AllReduce} communication primitive
combined with a parallel (online) learning
algorithm~\cite{Agarwal-terascale} which provides scaling
across cores and machines that is currently unnecessary.}

\xhdr{Offline Learner:} The Offline Learner allows for extensive offline
experimentation with the exploration data.  Training and evaluating
policies via alternative algorithms (such as those which cannot be
updated online, using different hyper-parameters, estimators
other than \ips, or different policy classes), trying out and tuning
new algorithms for exploration and policy learning, experimenting with
unused logged features, and switching to different reward
metrics are all possible.  The \mwt methodology guarantees offline
experimentation is counterfactually accurate~
\cite{Dudik-uai12,DR11,Langford-www10,Langford-wsdm11}.  Improvements generated through offline experimentation
can be integrated into the online loop by simply restarting the Online
Learner with the newly optimized policy or algorithm.

\ignore{\xhdr{Offline Learner:} Not all experimentation can be done online
in real-time so this module allows for extensive offline
experimentation with the exploration data.  Training and evaluating
policies via alternative algorithms (such as those which cannot be
updated online, possibly using different hyper-parameters, estimators
other than \ips, or different policy classes), trying out and tuning
new algorithms for exploration and policy learning, experimenting with
unused logged features, and switching to different reward
metrics are all possible.  The \mwt methodology guarantees offline
experimentation is counterfactually accurate~
\cite{Dudik-uai12,DR11,Langford-www10,Langford-wsdm11}.  Its fruits
are integrated into the online loop by simply restarting the Online
Learner with a newly optimized policy or algorithm.}

}

\subsection{Key techniques}
\label{sec:techniques}

\sid{Claim novelty more explicitly?}
We highlight several techniques that build on our abstractions
to further address our design goals.

\ignore{
\subsubsection{Robust logging and randomization}

Written and verified jointly by experts in systems, machine
learning, and software development, the Client Library consolidates
all randomization and logging logic in the application, which prevents many bugs
and simplifies further development of exploration capabilities.

\sid{move to client library discussion in arch subsec}
Most systems for machine learning assume reward information is
available at the same time as the interaction, which is rarely the
case in practice.  As previously noted in Section~\ref{sec:goals},
many bugs arise due to incorrect joining of this information. Our
design systematically eliminates such bugs by ensuring the $(x,a,p)$
tuple is set aside for logging at the point of decision and is
correctly joined with the reward later. This guarantees the feature
values used for policy evaluation are consistent with the interaction,
the correct probabilities are recorded and the action chosen by the
randomization is correctly logged even if downstream logic overrides
it.

Recording or using probabilities incorrectly has been the other key
cause of many past failures we have observed. The Client Library
implements randomization using a two-layer design: various exploration
policies sitting in the lower layer take as input a context and output
a distribution over actions; then the top layer samples randomly from
this distribution. Thus all randomization logic exists in one
place. Randomization occurs by seeding a pseudorandom number generator
(PRG) using the key $k$ and an application ID. The PRG is invoked
exactly once per interaction and is never reused across
interactions. Including the application ID in the seed ensures that
the randomization from multiple uses of the \ds in the same stack are
not correlated.
}

\subsubsection{Full reproducibility}

The \ds supports full offline reproducibility of online runs, despite
the presence of randomized exploration and continuously updated
policies. This addresses (F4).

The Client Library implements randomization using a two-layer design:
various exploration policies sitting in the lower layer take as input
a context and output a distribution over actions; then the top layer
samples randomly from this distribution. Thus all randomization logic
exists in one place. Randomization occurs by seeding a pseudorandom
number generator (PRG) with the key $k$ and an application ID; each
PRG is used for exactly one interaction.
Including the application ID in the seed ensures that the randomization from
multiple uses of the \ds are not correlated.

Policies trained by the Online Learner are stored with a unique ID in the
Store.  The Client Library records the ID of the policy used for each decision.
This, combined with our randomization scheme, ensures decisions are
reproducible.

The Online Learner may encounter reordered events from the Join
Service due to scale out, so it records the order in which interactions are
processed with the stored policies. Other learning events, such as periodic
resets of the learning rate to favor recent events (see \refsec{mwt}), are also
recorded. Together, this ensures each trained policy is reproducible.

We have not yet evaluated full reproducibility when scaling the Online
Learner as not even our largest deployments have required it.  In
principle this can be achieved by recording the parallel learning
configuration.  

\subsubsection{Real-time safeguards}
\label{sec:safeguards}

Each of our deployments has stressed the need for real-time monitoring
and safeguards against the unexpected. The \ds supports this,
addressing (F4).

\ignore{First, it is compatible with downstream business logic or
environmental constraints, an aspect of (M3).  Such logic may alter, bound,
or reverse the choices made by the \ds, for example to implement fail-safes
against violations of business policy or human safety.  Since the Client Library
correctly records the action taken at decision time, not any overriden values,
this does not affect the system's ability to train policies or provide accurate
counterfactual estimates of performance.}

The incremental nature of ips~(\ref{eq:ips}) ensures we can do policy
evaluation in real-time. The system supports this ability in the
Online Learner, which has a mechanism to specify candidate evaluation
policies, and the resulting estimates are displayed through a
dashboard (\reffig{arch}). This provides an accurate estimate of a
policy's performance before it is deployed, and also allows real-time
comparisons with default or ``safe'' policies. This monitoring is
novel and powerful because it is counterfactually unbiased, enabling
automated responses with confidence.  For example, one could roll back
a policy if it deviates too much from how the previous
best policy would have performed, or trigger alerts when the 
policy starts losing out to some reasonable baseline. More advanced
controls are possible, depending on the application.

\ignore{Defining what safety means for a learning system is challenging so the
automated safeguards provided by the \ds are powerful tools to enable
tricky goals.}

\subsubsection{Auto-generated features}
\label{sec:auto-features}

Through working with different customers, we observed that a large class of
applications---online content recommendation---benefited from the same kinds of
featurization. The Feature Generator makes it very easy for such
applications to use the \ds (see \reffig{api}). Any publicly accessible content
(\eg articles, videos, etc.) can be passed to the Feature Generator ahead of
time to be automatically featurized and cached---\eg text sentiment analysis,
video content indexing, etc.. These features are inserted on-the-fly by the
Client Library before making a prediction involving this content.

Our implementation of the Feature Generator (\refsec{impl}) does not
directly reduce the technical debt associated with input
dependencies~\cite{hitd}, because it relies on independent cloud services to
featurize the content. However, it shifts and consolidates the technical debt
from multiple customers to a single service (ours), which reduces costs in the
long term.

\subsubsection{Low-latency learning}
\label{sec:low-latency}

The \ds enables efficient, low-latency learning across its components,
addressing a key systems goal.

The Client Library supports multi-threaded predictions over a locally-stored
model, which is critical for front-end applications that serve many concurrent
requests (\eg \msn).  It also avoids sending repeated fragments of a context
(\eg features of an article that appears in many decisions) using a
simple encoding scheme: the fragments are sent explicitly only periodically,
and are replaced with references otherwise.
Our implementation has further optimizations (\refsec{impl}).

\ignore{
resulting in sub-millisecond parallel
decisions. It also optimizes context processing in two ways. In addition, we use
a simple caching scheme for repeated fragments of a context to substantially
reduce data transfer.  Such fragments are sent explicitly only periodically,
and are replaced with references otherwise. Our implementation has further
optimizations (\refsec{impl}).
}

The Join Service adds no fundamental delay on top of the experimental unit, but
due to the above encoding scheme, it may reorder events in a way that forces the
Online Learner to buffer references to features that have not yet been seen.
The Online Learner and Client Library use ML software supporting
millisecond training and sub-millisecond prediction.

\OMIT{As discussed in our \mwt methodology, the Online Learner optimizes a policy
via a reduction to supervised learning. By using an online algorithm for the
latter---currently we train a vector of linear weights---we can
incorporate new interactions extremely quickly and publish updated
policies for the Client Library to consume.  In addition, the Online Learner can
evaluate arbitrary policies in real-time, including the currently optimized
policy, by applying \refeq{ips} to the exploration data.  We use these
statistics to implement safeguards (\refsec{safeguards}) and to display
performance on a Power BI dashboard.}

All components support batching to improve throughput. Without
batching, the end-to-end latency of submitting an event and receiving a policy
that trained on it is \about{7} sec (\refsec{eval}).

\section{Implementation and Deployment Options}
\label{sec:impl}

\sid{If we need to save space we can trim a few places here.}

\begin{figure*}
\centering
\small
\begin{minipage}[t]{0.49\textwidth}

Sample interaction between \APP and web API to choose content:
\begin{enumerate}
\item \APP sends request to \ds:
\smallurl{https://ds.microsoft.com/api/decision/}\textbf{\texttt{\scriptsize{func}}}\smallurl{/APP/a1/a2/.../aN.js}\\
User/content features can be embedded manually (\eg
``\smallurl{/APP!location=NY.../a1;trending=3.2}'') or using Feature
Generator.
\item \DS responds with action: \smallurl{{Action:a2, EventId:X}} This
  is passed via JSONP to the \textbf{\texttt{\scriptsize{func}}} callback to
  render \smallurl{a2}.
\item \APP reports a reward (\eg click=1) using
  provided event ID:
  \smallurl{https://ds.microsoft.com/api/reward?reward=1&eventId=X}
Or, relies on a tracking pixel embedded into each content's page.
\end{enumerate}
\end{minipage}
\hfill
\rulesep
\hfill
\begin{minipage}[t]{0.49\textwidth}
Sample interaction between \APP and Client Library choose content:
\begin{lstlisting}[escapechar=\%]
var serviceConfig = new DecisionServiceConfiguration(settingsBlobUri: "<from deployment page>"); 
var service = DecisionService.Create<MyContext>(serviceConfig);
int action = service.ChooseAction(uniqueKey, myContext);
service.ReportReward(reward, uniqueKey);
\end{lstlisting}

The \algofont{ChooseAction} and \algofont{ReportReward} calls
correspond to steps 1 and 3 in the web API.  \algofont{MyContext} is any
class that has been annotated with JSON properties to identify
features, non-features, etc.~\cite{ds-clientlib}).
\end{minipage} 
\begin{minipage}[b]{1.0\textwidth}
\begin{tabularx}{\textwidth}{ X }
\\[-2ex]
\hline 
\end{tabularx}
\end{minipage}

\caption{The \ds API, accessible via a cross-platform web API (left)
or a low-latency Client Library (right).}
\label{fig:api}
\end{figure*}

Our implementation of the \ds is on Microsoft Azure~\cite{Azure}, but
our design is cloud-agnostic and could be migrated to another
provider. We describe noteworthy aspects of our implementation and the
deployment options. Our self-hosted service is open
source~\cite{ds-clientlib}.

The \ds presents a simple API, shown in \reffig{api}, implementing the
definition of a contextual decision from \refsec{mwt}. The API is
implemented by the Client Library or a web service that proxies
requests to it; the latter is built on Azure Web
Apps~\cite{azure-webapps}.  The Client Library links to Vowpal Wabbit
(\vw) ~\cite{VW} to predict locally on models trained by the Online
Learner, which also uses \vw. It is implemented in C\#, C++, and
Java. The C\# library is about 5K lines of code: 1.5K for exploration
algorithms ranging from \EpsGreedy to the theoretically-optimal
\algofont{Cover}~\cite{monster-icml14}, and 1.5K to handle batched
uploads to the Join Service.  We have since migrated the code for
exploration algorithms into \vw ($\sim$1K lines), ensuring consistent
parameters and logic between training and prediction.

The Join Service is implemented using Azure Stream Analytics (ASA)~\cite{ASA},
which allows us to specify a delayed streaming join using 45 lines of
query language.

The Online Learner is implemented in 3.2K lines of C\# as an Azure
worker role.  It uses \vw for online policy training and evaluation; we
modified \vw to support real-time evaluation of arbitrary policies (\ie \mwt
capability).  \vw reduces policy training to cost-sensitive classification, for
which many base algorithms are supported. We currently train a linear
model (a vector of weights), though decision trees, neural networks, and other
representations are available (\eg we are integrating CNTK~\cite{CNTK} into \vw
as a base learning algorithm).

The Store is implemented using Azure Blob storage. The Offline Learner schedules
jobs for policy training/evaluation on stored data using a distributed
analytics platform (we are currently migrating to Azure Data
Lake/U-SQL~\cite{azure-datalake}).

The Feature Generator may provide additional context for predictions. It uses
Microsoft Cognitive Services to featurize content from a URL and caches
the results, such as Text Analytics, Computer Vision, and Video
Breakdown.

Communication between the Client Library, Join Service, and Online Learner
(shown as arrows in \reffig{arch}) happens via JSON messages sent on queues
called Event Hubs.

\xhdr{Scalability and faults.}
Except for the Online Learner, all components can scale out and tolerate
faults using the Azure services they are built on.  The Event Hubs connecting
the components are also fault tolerant: they can replay data up to 7 days in the
past, which eases recovery.  
The Online Learner recovers from failures more coarsely: it loads the last
checkpointed model from the Store and replays the Event Hub data from that point onwards.

The Online Learner can scale out using \vw's
\algofont{AllReduce} communication primitive~\cite{Agarwal-terascale},
which enables parallel learning across \vw instances.

\xhdr{Optimizations for speed.}
To reduce the memory overhead of parallel predictions in the Client Library, we
modified \vw slightly to support sharing model state across \vw instances. Also, since the same
context class (\eg \algofont{UserContext} in \reffig{api}) is processed
repeatedly, we construct and reuse an abstract syntax tree to speed up 
serialization of the class during uploads to the Join Service. On the other
side, the Online Learner supports multi-threaded deserialization of joined
data to keep up with \vw's training (which is very fast).  

When using the web API, we leverage Azure FrontDoor's (AFD)
global edge network, which reduces client latency by maintaining persistent
connections. Additionally, the API allows contextual information to be
embedded---either manually or using the Feature Generator---in a
single background HTTP call (\reffig{api}). When called from a browser,
this allows the decision to manifest before the page finishes
loading.



\sid{How are policies specified for real-time evaluation?}

\ignore{
- Say built in azure at a high level
- summarize services used for some components. Say details for others below.
- Use event hub in between for fault tolerance. If the online learner fails,
just reload last model and replay events. 

- Details for client library, including optimization below and web api fast
deployment using AFD load balancing

- Online learner details

- Then deployment

- Figure for UI
}

\xhdr{Deployment options.}
The \ds supports push-button deployment (\about 6 min.) after a simple
registration~\cite{ds-portal}. Each option below has been
used for a real deployment.

\begin{noindentlist2}
\item {\bf Self-hosted:} This deploys a \ds loop in {\em your}
Azure account. This is ideal for customers like \msn who wish to 
control the deployment or limit exposure to data.

\item {\bf Hosted:} This deploys a \ds loop in {\em our} Azure
  account, allowing us to share resources across tenants. We omit
  discussion of our multi-tenant design, but the key elements are:
  distributing model state across the web API frontend servers and
  using AFD rules to direct each application's requests to the right
  server, creating a multi-tenant Join Service, 
  and consolidating Online Learners onto the same physical machines. 
  The hosted service is being used by \complex.

\item {\bf Local mode:} This deploys a \ds loop locally in
your machine using extended functionality in the Client Library.
The Join Service is replaced by an in-memory buffered join, and models are
trained and invoked for prediction by local VW instances. Local mode is
particularly useful for testing in simulated environments (\eg network/cloud
simulators). It is being used by \ac.

\end{noindentlist2}

\section{Deployments}
\label{sec:deployment}

We describe five applications of the \ds across diverse domains, as
represented in \reftbl{apps}. We focus on problem framing and
deployment characteristics here; evaluations and lessons learned are discussed in \refsec{eval}.

\begin{figure}
\centering
\includegraphics[scale=0.19]{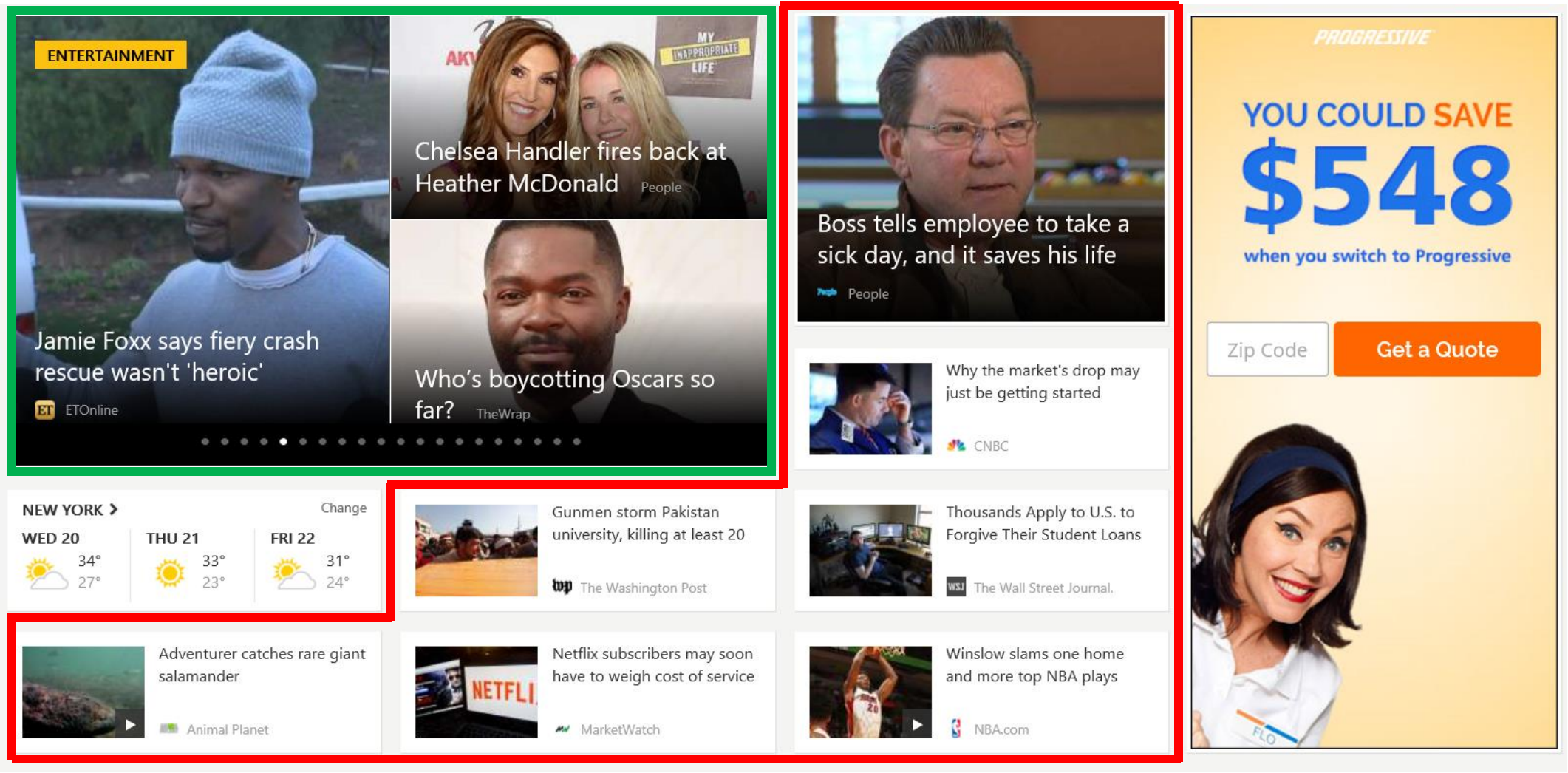}
\caption{\msn homepage. The Slate (boxed red) and Panel (boxed green)
use the \ds in production as of early 2016. Ads to the
right (not optimized) also saw a lift.}
\label{fig:msn}
\end{figure}

\xhdr{\msn.}
The \ds personalizes news stories displayed on the \msn homepage shown in
\reffig{msn}.  The deployment is now the production default, serving
10s of millions of users who issue thousands of requests per second.

\msn's problem is essentially the \news problem from \reftbl{apps}: a user
requests the homepage and \msn's front-end servers must
decide how to order the articles on the page.  If a user is logged in, there is context:
demographics (\eg age, location) and the topics of news stories they
have clicked on in the past; otherwise only
location is available. The action choices are the current set of news
articles selected and ranked by editors (tens of articles,
typically). Each article has features that describe its topic. The
reward signal is clicks.

The \ds optimizes the click-through rate (CTR) on the
article in the most prominent slot of a segment, unless it is locked by the
editors, and uses the resulting policy to pick articles for all
slots.\footnote{There are approaches for optimizing all slots simultaneously~\cite{AkshayCB-2015}.} The
default exploration policy \EpsGreedy is used with $\eps = 33\%$.
The experimental unit is set to 10 minutes, and a new model is
deployed to the Client Library every 5 minutes. These settings are
driven by the data rate and how non-stationary the news
application is. \ignore{For example, an initial experiment showed poor
  performance due to a combination of low click rate and a
  click/no-click reward encoding of $\{-1,0\}$ instead of $\{0,1\}$.
  This seemingly minor difference had a huge impact on the variance of
  estimators.  Since then, we have used the new encoding in all our
  deployments and have modified the core algorithms in \vw to better
  compensate for encoding errors.}

Before using the \ds in production, \msn used standard A/B testing to compare
its performance with the editorial ordering, the default policy which beat
previous attempts to use machine learning.  Experiments on the Slate segment
of the page showed a >25\% CTR improvement over a two-week period; the Panel
showed a 5.3\% improvement, which is even more significant as it receives 10x
more traffic than the Slate.  These gains were achieved while maintaining or
improving long-term engagement metrics such as sessions per unique user and
average session length, showing that the CTR metric is aligned with longer-term
goals in this case. 

The success of the experiments led \msn to make it the default in
production for all logged-in users in early 2016.  The \msn team has
since deployed the \ds a half dozen ways. \msn's deployments are
self-hosted and they have customized some of our components. For
example, they implemented a multi-tenant Join Service using Redis
Cluster~\cite{redis-cluster} that is shared across applications (\eg
different page segments). Our modular architecture allowed this
customization, while ensuring correct semantics were preserved.  \msn
uses the C\# Client Library in their front-end servers for
low-latency, but clicks are reported directly from user web browsers
via the web API.  A single Online Learner is used for each
application to isolate the trained models.

\xhdr{\complex.}
\complex wanted to use the \ds in two ways: to
recommend videos to associate with news articles and to recommend top
news articles.  These applications are similar to \msn but with
several significant variations:
\begin{enumerate}
  \item The amount of traffic is much lower, $\sim$ 35K events/day.
  \item Features from \complex are much sparser, but
    high-dimensional, derived from keywords or other metadata.
  \item \complex uses a content distribution network implying that
    all significant computation must be offloaded.
\end{enumerate}
The first two variations make application of the \ds much
more statistically sensitive than with \msn.  The third variation
forces the use of a cloud API (Fig.~\ref{fig:api}, left). We leveraged
our feature generators to enhance \complex's features with more
informantive ones. The \ds provide a lift of >30\% over
editorial.

\xhdr{\tr.} \tr optimizes landing pages for in-app
advertisements so as to maximize revenue with event rates of around
200k/day.  This varies from the previous two content recommendation
applications because the revenue signal is scalar rather than binary
and much sparser.  Request rates also vary enormously depending on the
campaign.  The baseline here is an $\epsilon$-greedy bandit algorithm,
that does not incorporate context. This baseline is structurally
similar to our approach in this deployment besides the use of context,
so the 18\% lift provides a measure of the value of context.

\xhdr{\toronto.} \toronto wants to reduce technical support
load by automatically providing answers to technical support
questions.  There is a large negative reward associated with customers
requesting human intervention.  This problem significantly differs
from the previous in the context (technical support queries), the
nature of the actions (pointers to solutions), a low event rate, and
privacy concerns. Fortunately, \toronto has existing systems and
expertise winnowing down the set of actions to a reasonable set and
good actions are stable over time allowing us to handle the low data
rate. Thee privacy concerns are addressed by simply deploying the
service into their own account.  Results are too preliminary to
report.

\xhdr{\ac.}  \ac wants to contextually optimize dealing with
nonresponsive virtual machines in a cloud service.  In particular, if
you migrate too soon the cost of migration may exceed the cost of
waiting for the VM to become responsive.  Alternatively, waiting for a
VM that never becomes responsive is a pure waste of time.  A very
large number of sparse features are available related to the machine,
the VM, and the process on the VM.

Unlike previous applications a large effectively supervised dataset
exists because the current baseline policy in their system is to
simply wait for 10 minutes, making it possible to simulate the effect
of any action involving migration before 10 minutes. Taking such a
dataset collected with this baseline, we estimated a 19\% reduction in
wasted time, significant enough to proceed to production. Note that
once the system is deployed, we can no longer evaluate all possible
options for free, as the system may choose to immediately migrate, and
the \mwt capability of the \ds is critical to computing real-world
performance estimates.

\section{Evaluation and Lessons}
\label{sec:eval}

We evaluate the \ds based on how well it meets our systems and technical debt
minimization goals (\refsec{motivation},\refsec{goals}). We use both live deployments and the
exploration data collected from them.

\subsection{System evaluation}
\label{sec:system-eval}

\sid{How do we address comparisons to other familiar systems, such as
Tensorflow serving or Clipper? We should say either here or at top of Related
work that the Decision Service is the first system of its kind.}

From a systems perspective, we answer the following:
\begin{enumerate}
  \item What is the latency and overhead of making decisions? How 
  quickly is data incorporated into trained policies?
  \item Can the system scale to high event rates?
  \item Is there adequate support for offline experimentation?
\end{enumerate}

\xhdr{Experimental methodology.} We deployed a self-hosted \ds loop
and drove traffic to it using the sample code distributed
with the Client Library~\cite{ds-clientlib}, running on several large A4
instances (8 cores, 14GB memory, 1Gbps NIC).  Except the Online Learner, all components including the
Event Hubs connecting them can be scaled by configuring the underlying Azure
service. The Online Learner is a stand-alone worker running on a D5
instance (16 cores, 56GB memory).

Some of our experiments use exploration data collected from the \msn and
\complex production deployments during April 2016 and April 2017,
respectively.
This data contains the real $(x,a,p,r)$ tuples generated on those days with $x$ consisting
of \about 1K features per action.

\subsubsection{Latency of decisions and learning}

We are interested in both \emph{decision latency} and \emph
{learning latency}.  Decision latency is the time to make a decision in the
Client Library (\ie the \algofont{ChooseAction} call). Learning latency is the
time from when an interaction is complete (\ie \algofont{ReportReward} has been
called) to when it affects a deployed policy in the Client Library.

We measured the decision latency by training a policy on one hour of
\msn data, deploying this policy in the Client Library, and then
repeatedly calling \algofont{ChooseAction}. The average latency
is 0.2ms, well within the needs of our customers. Latency is not
the only metric that matters, however. For example, \msn's front-end servers are
CPU limited, so CPU/req is monitored very carefully. The
\ds increased CPU/req by 4.9\%, which was deemed acceptable.

\begin{figure}
\centering
\includegraphics[scale=0.42]{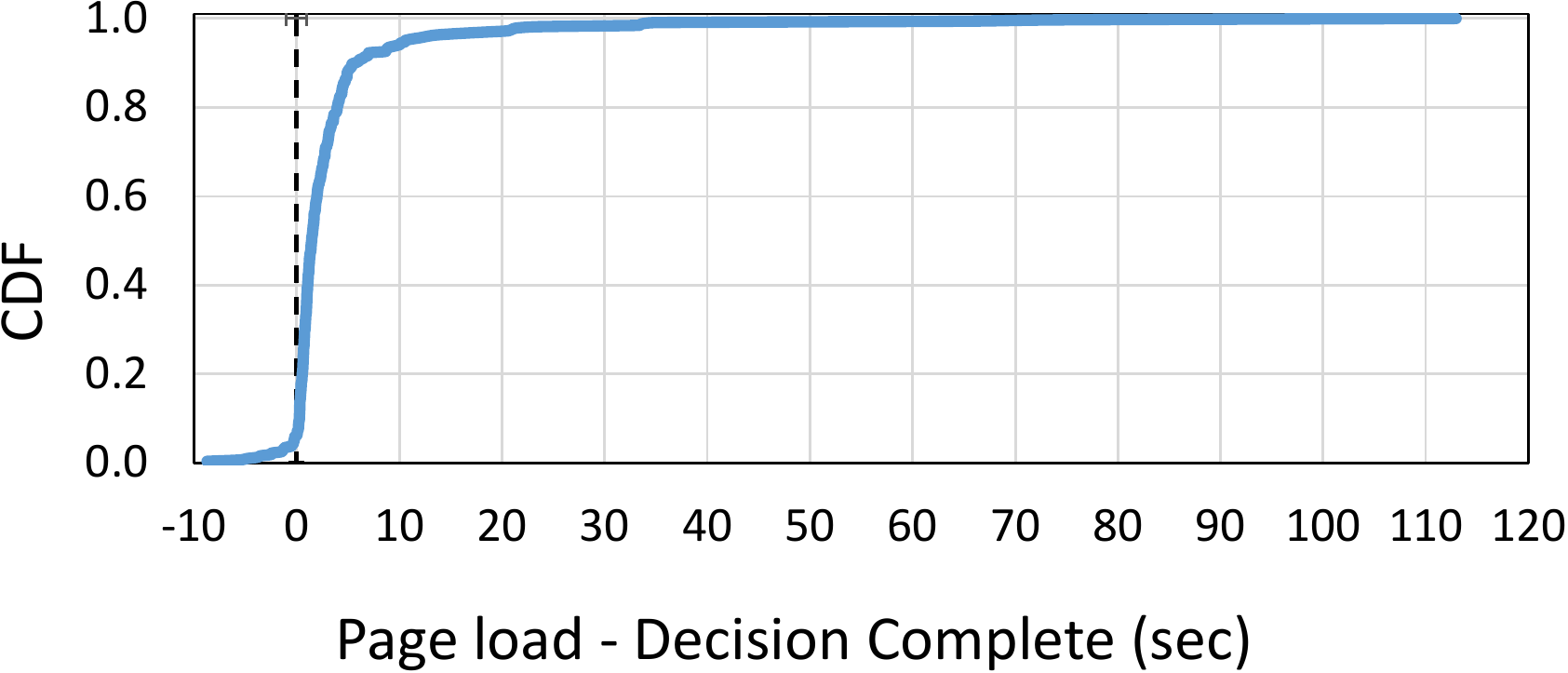}
\caption{Decisions complete before page loads in \complex.}
\label{fig:browser-latency}
\end{figure}

In our implementation, we described an optimization that leverages Azure's
edge network and JSONP calls to our web API to make fast decisions from
a browser. This is enabled in the hosted \ds used by \complex.
Using page load metrics across 20 days, we measured the difference
between page load time and decision time (steps 1 and 2 in \reffig{api}) for
\complex. \reffig{browser-latency} shows the results: except for 5.9\% of
requests, all decisions complete before the page finishes loading, and less than
1\% take >4 more seconds (the max is 8.7sec).

To measure learning latency, we first removed any configurable sources
of delay, such as batching and caching (which could cause downstream
reordering). We configured the Client Library to poll for new models every
100ms, we set the Join Service experimental unit to 1 second, and we
configured the Online Learner to publish an updated policy after each event.  We
then replayed one event from the \msn data and waited for a policy to appear.  The
average learning latency is 7.3 sec. 


\subsubsection{High event rates}

\ignore{
\begin{figure}
\centering
\includegraphics[scale=0.27]{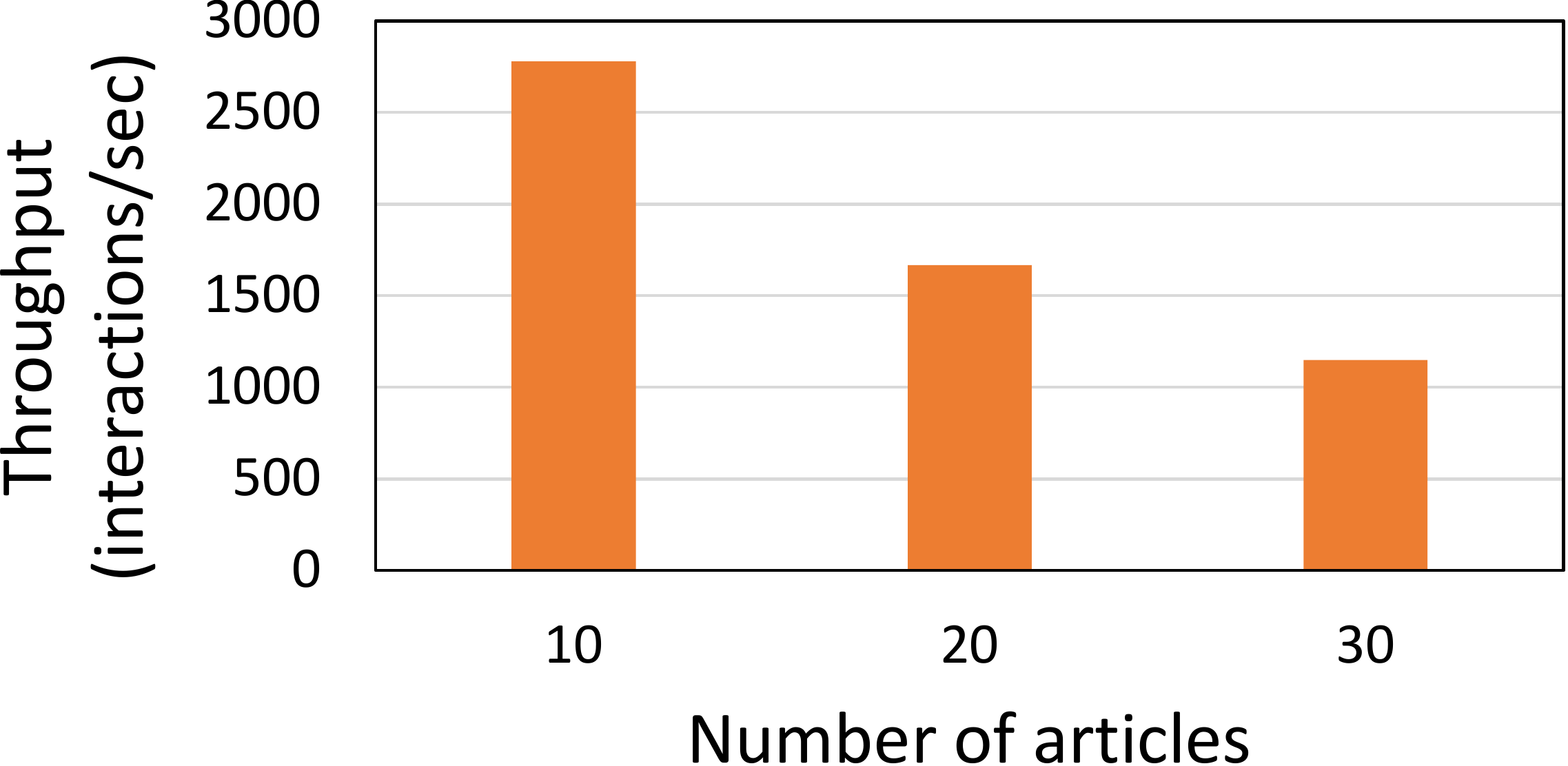}
\caption{Online Learner throughput for different article counts per
interaction.}
\label{fig:vw-actions}
\end{figure}
}

Most of our components are built on scalable services managed by
Azure. The Feature Generator relies on Microsoft Cognitive Services
APIs, which are also scalable (though outside our control). Thus we
focus on the Online Learner.

The data rates seen in all our production deployments have been
adequately handled by learning on a single core. To saturate the Online Learner, we
preloaded \msn data into the Join Service's output Event Hub and processed
it at full speed. \msn uses the Client Library's encoding scheme (\refsec{low-latency})
to reduce data size. The throughput achieved by the
learner was stable at 2000 events/sec, implying 100 million events/day
applications are viable.  Buffering for events whose encoded features  had not
arrived was minimal, remaining at 0 most of the time with occasional spikes up
to 2250 events.

We observed two different bottlenecks in the Online Learner. When using encoded
contexts (\eg \msn), the smaller data makes policy training (\vw) the
bottleneck.  For example, increasing the number of articles (actions) per
decision in \msn from 10, 20, 30---which slows down training without
increasing data size too much---led to a throughput decline of 2778,
1666, 1149 events/sec.
When contexts are not encoded (\eg \complex), the bottleneck is the
deserialization of data received from the Join Service. We verified this by
changing the \vw parameters to slow down training without seeing any decline in
throughput, at \sid{XX} event/sec.

Although we have been able to scale the deserialization overhead near-linearly
using multiple Online Learners, we have not yet incorporated \vw's
parallel learning so cannot comment on model performance. We plan to address
this when the need arises and do not expect new insights over prior
work~\cite{Agarwal-terascale}.

\subsubsection{Offline experimentation}

The Offline Learner has enabled us to run hundreds of experiments on
exploration data, across our deployments, to tune parameters, try new
learning algorithms, etc.. For example, \complex began with a simple
\EpsGreedy policy to collect initial data. Using this data, we ran 
parallel offline experiments and discovered higher CTRs when using an
ensemble exploration algorithm called
\algofont{Bag}~\cite{monster-icml14}, a different learning rate, and
certain feature interactions and omitted features. This was determined without
any additional live experiment, and our ML methodology guaranteed that the
results were counterfactually accurate.

\ignore{
\sid{Cut if need to save space}
Offline experiments can also reveal the value of features.
For example, roughly half of the CTR lift in \msn is due to 
users' demographics and the other half to article topic features.
}

\begin{figure}
\centering
\includegraphics[scale=0.48]{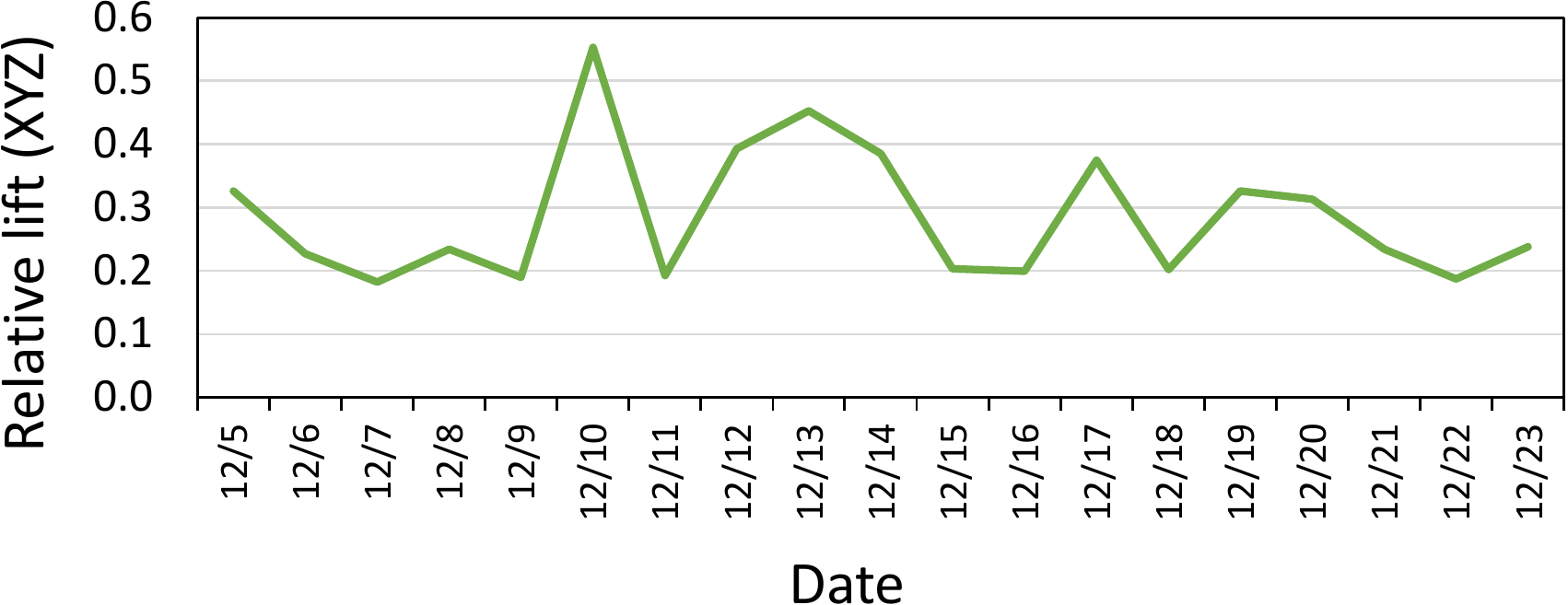}
\includegraphics[scale=0.48]{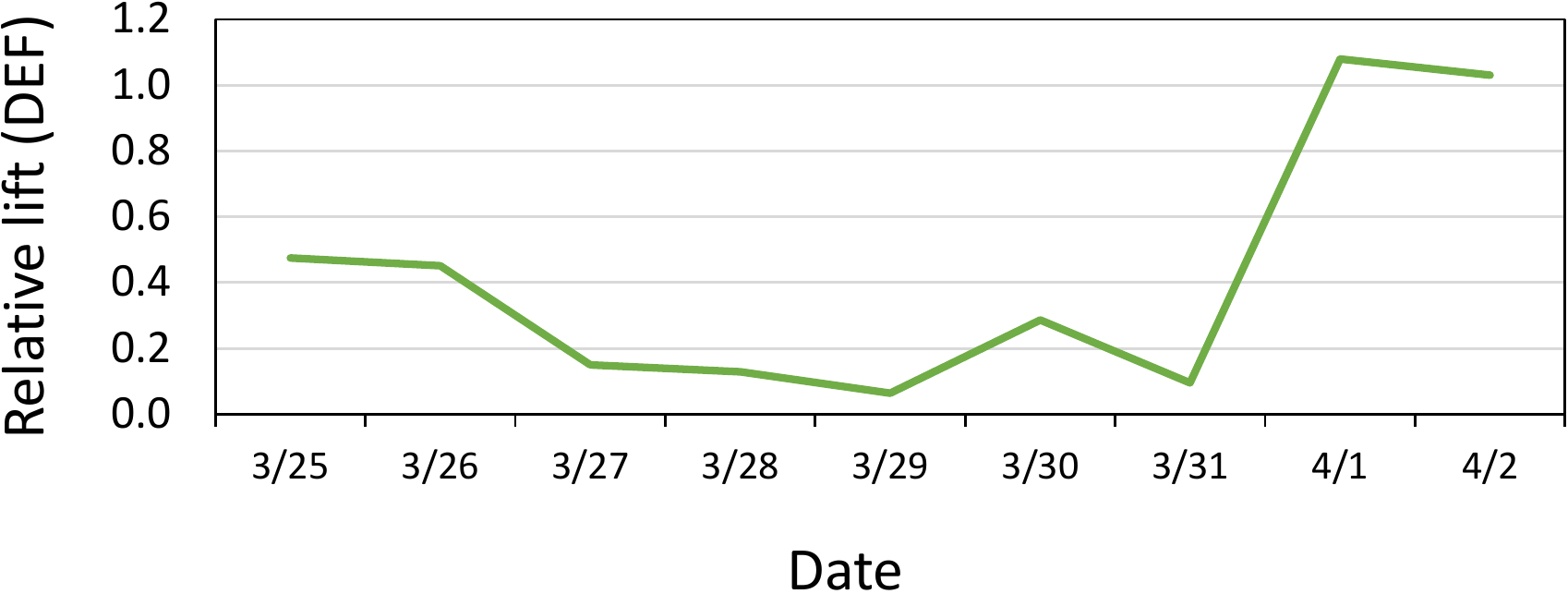}
\caption{Daily CTR lift for \msn flight (Dec. 2015) and \complex flight (Mar.
2016).}
\label{fig:ctr}
\end{figure}

\subsection{Technical debt evaluation}
\label{sec:techdebt-eval}

From the perspective of technial debt, we ask how well the \ds addresses
problems F1-F4. As mentioned in \refsec{motivation}, a common theme across
the problems is ensuring that performance predicted at learning time matches
observed performance at decision time. Thus, most of our experiments
measure the discrepancy between the two caused by failures.

\ignore{
\begin{enumerate}
  \item Are policy evaluation estimates reliable? Can policies be
    compared in real-time?
  \item How important is it to continuously train the policy?
  \item What effect does bad logging have on the policy?
\end{enumerate}
}

\xhdr{Experimental methodology.}
To simulate failures on exploration data, we take a day's data and randomly
allocate 80\% of it for training and 20\% for testing while preserving time
order, approximating the normal train/test split methodology of supervised
learning. We modify the training data with the failure and train a policy
on both the faulty and correct data; this yields an estimate of performance
via the policy evaluation capability of the Online Learner (based on
\refeq{ips}). We then evaluate both policies on the test data and obtain
another estimate of performance. We report the discrepancy between the training
and testing estimates, which for the correct data is always within 5\%.


\ignore{
The reason we did not split a day's data to the first 80\% and the last 20\%
is because we observed substantial intraday nonstationarity in both \msn and
\complex. This made it difficult get consistent results between training and
testing using the baseline (correct) data. By randomizing the train-test split
over the entire day (but preserving time order), we obtained consistent
performance (within 5\%).
}

\subsubsection{(F1) Partial feedback and bias}
\label{sec:eval:bias}

By using contextual bandit exploration, the \ds avoids the bias
inherent in approaches like supervised learning. This is evidenced by the
superior performance it achieves compared to the editorial ranking in
both \msn and \complex, shown in \reffig{ctr}. Not only are the per-day
improvements high (between 18\% and 55\% for \msn, between 7\% and 112\% for
\complex), but the improvement is maintained over time (see F3 below).

\ignore{
It is a natural urge to want to shield the most critical and high-value portions
of an application from exploration.  For example, the \msn editors periodically
locked important slots from the recommendations of the \ds. While this
limits the downside, it also limits the upside as we can only optimize less
impactful portions of the page.  In our deployment, we found each time the
editors unlocked a more important slot, the CTR and engagement overall improved significantly.
}
 


\sid{add reward delay fig?}

Another form of bias occurs if rewards for some actions are delayed more
than others. We simulated this effect in the \msn data by moving the click
events of certain popular articles before those of the rest, as if the Join
Service had immediately emitted them instead of enforcing a uniform experimental
unit.  As the Online Learner incorporates new data, the trained policy may
forget the importance of these articles. Indeed, shifting the
clicks of the top 2 articles yields a training performance that is 1.3x higher
than test performance (\reftbl{data-failures}).  If a constant
learning rate is used (as in \complex), this increases to 1.7x.

\ignore{
We used a constant step size for training as would be the case in some deployments (\eg \complex), which means that the learning algorithm would eventually forget about the importance of the shifted articles.  
We then used our train-test methodology above to evaluate the
consistency between training and testing performance. The result is shown in
\reftbl{data-failures}. Whereas the unmodified data yields consistent results,
the modified data with just 2 articles shifted yields test performance that is
1.3x lower than train performance with \msn's step size scheme, and 1.7x lower
with \complex's step size scheme.
}


\subsubsection{(F2) Incorrect data collection}
\label{sec:eval:data}

\begin{table}
\centering
{\footnotesize
\begin{tabular}{|l|c|}
\hline
{\bf Failure} & {\bf Train/test perf. discrepancy}
\\
\hline
Reward delay bias & 1.3x \\
\hline
Incorrect probability & 3.0x \\
\hline
Decision as feature & 8.7x \\
\hline
Modified feature & 1.2x \\
\hline
Deleted feature & 2.4x \\
\hline
\end{tabular}
}
\caption{Various data collection errors
(\refsec{eval:bias},\refsec{eval:data}).}
\label{tbl:data-failures}
\vspace{-2pt}
\end{table}

We evaluated four different data collection failures using \msn data and our
described methodology, all of which we have experienced in practice.
\reftbl{data-failures} summarizes the results.

\begin{noindentlist2}
\item {\em Incorrect probability.} A common error is when editors or
  business logic override the chosen action and record the override
  making the recorded probability incorrect.  We simulated this by
  overriding 10\% of the training data actions.

\item {\em Decision as feature.} Another common error is when the identity
or probability of the chosen action is used as a feature for downstream learning. 
We simulated this by adding a new feature to the training data that is 1 for
the chosen action and 0 for all other actions. Note that this feature would
not appear in test data because it is only available after decision.  

\item {\em Modified feature.} Features such as the user browsing history in \msn
are often modified by separate processes, and hence may change
between decision and learning time. We simulated this by replacing the browsing
history in 20\% of the training data with default values, \eg as if they were
still being computed for a new user.  
 


\item {\em Deleted feature.} A deleted feature (\eg due to a
  database failure) may be present at learning but not at 
  decision time. We simulated this by removing user demographic
  features.
\end{noindentlist2}

In all of the above cases, the estimate of the policy's performance
during training deviated from its performance during
testing, by a factor ranging from 1.2x to 8.7x. This undermines the
entire value of a counterfactually accurate system.
By correctly logging at the point of decision, the \ds avoids these failures. 

\subsubsection{(F3) Changes in the environment}
\vspace{-1pt}

We observed significant non-stationarity in the \msn and \complex data,
which is likely due to the continuous arrival of new content and user interests
swaying to breaking news or events.  The \ds is able to sustain its improvement
in \reffig{ctr} by continuously learning {\em and} periodically adjusting its
learning rate to favor recent data, as prescribed by our ML methodology.
Without this, performance would degrade over time.

Periodic resets of the learning rate---\eg in all \msn deployments the learning
rate is reset each day---may affect the Online Learner's ability to converge to a good policy
quickly.  To investigate this, we played a full day of \msn data and tracked the
performance of the trained policy relative to the editorial policy
(Editorial 1 in \reffig{df}). Both the trained and editorial
policies exhibit high variance initially, but the estimates become statistically significant with more data.
The trained policy  starts outperforming editorial after just 65K
interactions---for a request rate of 1000/sec, this is about 1 minute---and
eventually achieves a 42\% improvement by end of day.

To demonstrate the cost of not continuously training, we took three
days of \msn data, trained a policy on day 1 and tested it on days 2 and 3
(without updates). The performance relative to a policy trained on the corresponding
day is low:
\vspace{-4pt}
\begin{center}
{\scriptsize
\begin{tabular}{|l|l|l|l|}
\hline
Policy from: & Day 1 & Day 2 & Day 3 
\\
\hline
Same day & 1.0 & 1.0 & 1.0 \\
\hline
Day 1 & 1.0 & 0.73 & 0.46 \\
\hline
\end{tabular}
}
\end{center}
\vspace{-4pt}

In other words, day 1's policy achieves 73\% of the CTR of day 2's policy when
tested on day 2, and 46\% of day 3's policy on day 3.  This suggests that the environment and
articles have changed, and day 1's policy is stale. 

Some of the data collection errors in the previous section are caused by changes
in how features are generated. For example, the baseline policy in \tr uses
click statistics that are aggregated with a certain decay.  When we tried to use
\vw's built-in marginal statistics to match this feature, it took several weeks
to obtain consistent results.  This was partly because \tr's method of
collecting the statistics changed over time, and different details were
conveyed to us at different points. Continuous learning can help cope with such
uncertainties.


\subsubsection{(F4) Weak monitoring and debugging}
\label{sec:t4-eval}

\begin{figure}
\centering
\includegraphics[scale=0.38]{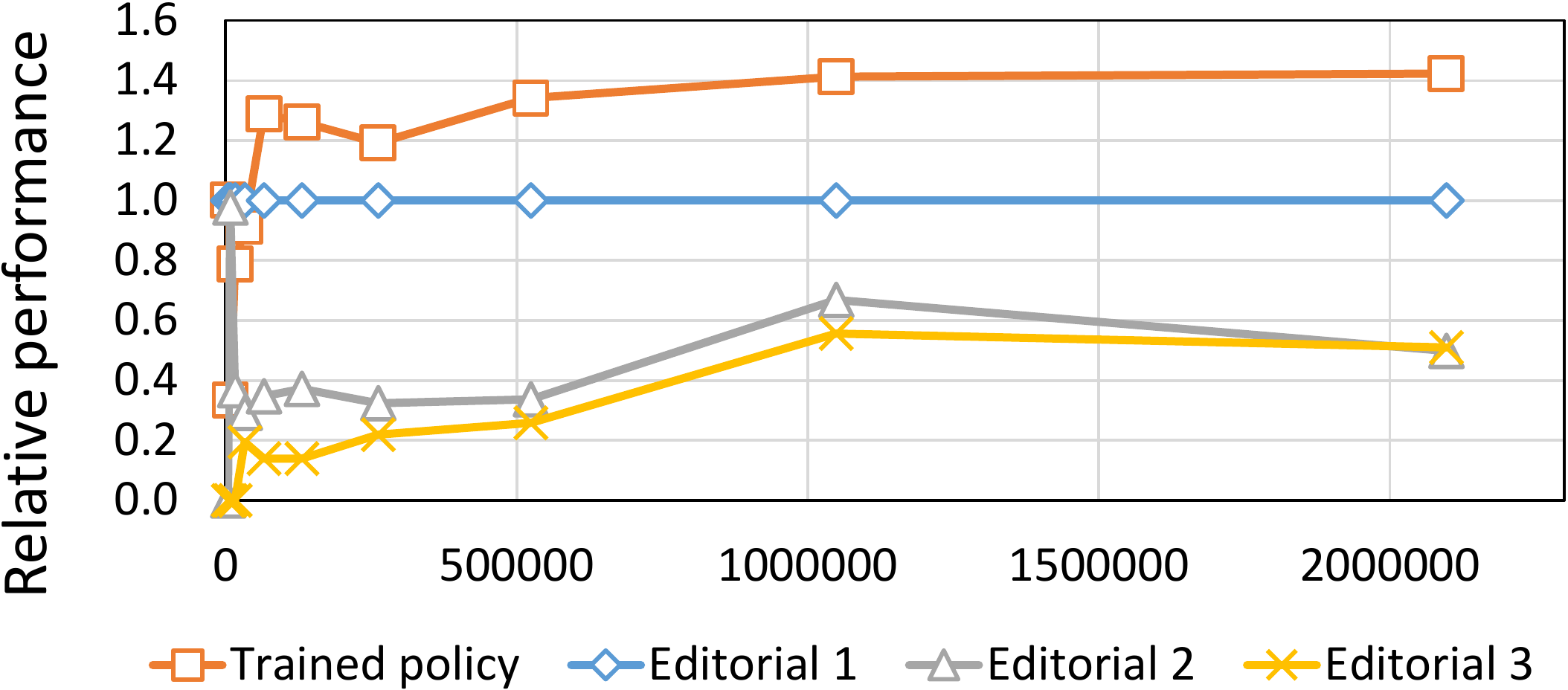}
\caption{Real-time comparison of trained policy against editorial policies,
normalized to Editorial 1's performance.}
\label{fig:df}
\end{figure}

The \ds supports real-time evaluation of arbitrary policies, an extremely
powerful capability. To demonstrate this, we took 3 simple policies which always
choose the first, second and third article respectively from the editorial ranking in \msn. The first
policy corresponds to the editorial baseline, while others are reasonable
alternatives. \reffig{df} shows a real-time comparison of policy performance
generated by a running instance of the \ds that we replayed \msn data
through. Each datapoint was captured within an average of 10 sec from when
the corresponding interaction completed. Although not in production yet, several
of our deployments have plans to install alerts based on these estimates. 

\ignore{
Separately, we have verified the accuracy of
these estimates compared to a control flight with very high precision. Across
all policies and days, the relative difference between the \ips and control
values was no more than 2.5\%, and all \ips estimates were within a 95\%
confidence interval around the control values.
}

\ignore{
While policy estimation using \ips as per~\refeq{ips} has theoretical
guarantees, it is prudent to verify this empirically. We took 3 simple
policies, which always choose the first, second and third article
respectively from the editorial ranking for the top slot. The first
policy corresponds to the editorial baseline, while others are
reasonable alternatives. For each policy, we computed an \ips estimate
of its value from a day's worth of data for two different days in
April. Additionally, we estimated the performance of each policy in
the control flight with a very high precision. Across all policies and
days, the relative difference between the \ips and control values was
no more than 2.5\%, and all \ips estimates were within a 95\%
confidence interval around the control values.
}

\ignore{
Having real-time estimates of arbitrary policy performance, such as baseline or
known ``safe'' policies, can enable powerful safeguarding logic that reacts to
violations in policy or safety in real-time. We do not prescribe such a policy
here---it depends on the application and is outside our scope---but note that
the \ds enables it.
}

The \ds supports full offline reproducibility of online runs by design. We
cannot over-emphasize the value of this property, as it has helped us diagnose 
numerous failures and performance issues in the course of our deployments. 
Several of these failures were diagnosed by simply detecting the point at which
the offline run deviated from the online run, \eg revealing that data was
dropped. However, the real value came in diagnosing problems in the ML software (\vw).
Typically such bugs are difficult to diagnose because the ML software is treated
as black-box, but with full reproducibility we can quickly rule out any of our
components as being the culprit.  For example, this helped us debug a poorly
performing \msn model that was initialized from another model trained
with an insufficient range of data. This triggered a bug in \vw that
inadvertently reset the min/max prediction range in a way that was incompatible
with the actual rewards of $\{-1,0\}$.

\ignore{
Essentially, a rogue parameter was added to the saved model that
triggered a null update on minimum and maximum prediction values, which meant that all of the updates were based on a prediction of 0.
Since some prediction should be less than 0 (we are using \{-1,0\} costs with
\msn), this meant that the system kept trying to drive the predictions smaller
without bound. Surprisingly, this kind of works, but it fails with articles of
different ages.
} 

\ignore{
As with any estimator, policy evaluation has variance that may be high or
low depending on circumstances. For example, we noticed a discrepancy with
our offline performance predictions for \tr, where we were predicting large
lifts but then seeing none after deployment. We realized we needed more robust
policy evaluation techniques that incorporate the exploration done during
deployment, tolerate long reward delays, and non-stationary policies which are
much more difficult to evaluate. Since then, our offline estimates have
generally been accurate predictions of online performance, which has enabled
several flights or deployments to go through with our customers.
}

\ignore{
\subsection{Continuous policy training}

To demonstrate the importance of continuous training, we used the policy
from the previous experiment (trained on day 1) and tested it on day 2 and day
3. We compared this to a policy trained on the corresponding day, and recorded relative performance:
{\small
\begin{center}
\begin{tabular}{|l|l|l|l|}
\hline
Policy from: & Day 1 & Day 2 & Day 3 
\\
\hline
Same day & 1.0 & 1.0 & 1.0 \\
\hline
Day 1 & 1.0 & 0.73 & 0.46 \\
\hline
\end{tabular}
\end{center}
}
In other words, day 1's policy achieves 73\% of the CTR of day 2's policy when
tested on day 2, and 46\% of day 3's policy on day 3.  This suggests that the environment and
articles have changed, and day 1's policy is stale. Continuous training solves
this problem.
}

\ignore{
\subsection{Data collection failures}

Our motivation for building the \ds came from witnessing past failures
to apply \mwt.  Many of these failures were due to incorrect
exploration data. We simulated two failure modes using the offline
logs from \msn. For each experiment, we took a day's data and randomly
allocated 80\% of it for training and 20\% for testing while
preserving time order, approximating the normal train/test split
methodology of supervised learning.

The \ds using the training set simulates the policy learned in
production, while the test set simulates the performance of this
policy when deployed online.  This simulation is imperfect because
training occurs on fewer events and because the policy evaluated on
the test set is not adaptive.  When training on the train set
progressive validation techniques~\cite{progressive} allow an unbiased
estimate of policy value deployed with zero delay.  When
evaluating the final trained policy on the test set, only a 0.9
fraction of performance is observed, implying that intraday
nonstationarity is important.

We simulated incorrect logging of probabilities which could be caused
by editors overriding 10\% of the actions and recording the override
rather than chosen action.  We trained a policy on the training set
and evaluated it on the test set. The progressive validation CTR from
training was 3 times higher than performance on the test set.  Hence,
the offline estimates of a policy's quality carry little meaning once
the probabilities are corrupted.

The second experiment simulates a common error whereby the identity or
probability of the random action chosen by exploration might be used
as a feature for downstream learning. We simulated this by adding a
new feature to our training data which was 1 for the action sampled by
\ds in our logs and 0 for all the other actions. The test data was not
augmented with this feature since the random choice is not available
to the policy at prediction time, but is rather based on the policy's
prediction.  We found the progressive validation CTR was nearly
  8.7 times larger than that observed on test data, again making the
  offline estimates meaningless.
}

\ignore{
\begin{enumerate}
  \item Does the \ds provide low latency decisions? Are interaction tuples
  incorporated into the learned model with low latency? \\
  - Use MSN scenario as setting. \\
  - Function of model size and blob upload rate \\
  - Use open source multiclass supervised learning dataset to drive this
  \item How does window size relate to throughput?  
  \item Does the \ds scale to meet the throughput demands of the application? \\
  - Report throughput single training machine and scaling out join server.
  \item Does the \ds recover from failures in any of its components?
  \item Does the \ds provide accurate counterfactual estimates of online
  policy performance? (Might be covered in deployment experiences)
  \item Is the \ds easy to setup and use?
  \item How does failure to do MWT correctly affect the system? \\
  - Set epsilon to 0 \\
  - Feature corruption \\
  - Not recording probabilities, recording the wrong probability \\
  - References to a database for features, either swaping out wrong features or
  adding noise \\
  \item Can the \ds be used to safeguard a system in real-time?
  \item What is the cost of deploying this on the cloud? Price per decision.
\end{enumerate}

Machine learning questions:

\begin{enumerate}
  \item Use offline eval to compare different policies (blend with above)  
  \item Do we need to compare model performance on an open source data set?
  General preference is to not go down this rathole and instead point to MSN
  data.
\end{enumerate}

MSN questions: 

\begin{enumerate}
  \item Some numbers mapping their performance to our microbenchmarks
  (throughput, latency, etc.). Ask for other system performance numbers
  \item Did the \ds outperform editorial ranking of articles?
  \item Did the improvement cannibalize clicks from other parts of the page, or
  affect health metrics?
  \item MUID vs ANIDs
  \item Infopane
  \item How much of the improvement came from context features?
  \item Can we show CTR going up over time according to MSN periodicity, show it
  against offline eval theory
\end{enumerate}

Plots needed:

\noindent Microbenchmarks:

\noindent System Performance: \\
-Choose Action Latency -client; Choose Action Latency -web api \\
-Complete Loop Latency - count each component's contribution \\
-Maximum Scale/throughput for a single training service \\
-Price per choose action - replay MSN data and calculate costs per our data or
calculate the cost based on a maxed out single training service \\ 
-Vary experimental unit duration and show cost/performance (throughput of JS
mainly) \\
--- In addition, should we do separate throughput experiment for JS, or is this
obvious? \\
-Failures: for JS just say its managed or lose data if using in-memory. Then
focus on online trainer failure, which can be restarted (how do we show the
effect of the restart, i.e. some sort of ``blip'') \\
-Easy setup: probably only thing to measure is time for push button deployment
to complete.

\noindent Learning Performance: \\
-Safeguards and real-time offline evaluation. Show we can do it by comparing
model to constant editorial policies. \\
-Show speed of learning by comparing model's performance as data goes from 0
up to final editorial model's performance (all using offline eval). This should
be complemented by MSN's ``start of flight'' data below, if we can get it. \\
-\sscomment{Not sure about this one} Offline Eval - compare a bunch of
different models, cite ML papers,

\noindent Tests our failure claims: \\
-Set epsilon to 0 \\
-Double the best class's probability \\ 
-Injecting noise into the features (swap features or simulate staleness) \\
---Using probability as feature?

\noindent MSN Data:

\noindent System Performance: \\
- \#rate of interactions and whatever else we can get from MSN folks to map to
our microbenchmarks, or show the delta due to their optimizations. \\
- Join server is multitenant can share with many apps. If you have 3000 req/sec
but one app, then latency suffers. If have three apps, separate key,value entry
for each timeslice. Each timeslice they use is 30 seconds, so this is the
issue.\\
---Kept latency at real-time and measure max throughput while varying number of
apps, and number of frontend servers (webapi), since frontend servers doing all
the real work
\\u
- VW online learner's performance. Main issue is number of documents that
affects events/sec. When you increase number of documents, VW becomes the
bottleneck. The expansion of doc vectors is done quickly (cached in memory), the
bottleneck is VW (unclear exactly what).

\noindent Learning Performance: \\
-CTR uplift over editorial ranking: real A/B test data. Breakdown into: \\
---Value of personalization \\
---Value of optimization \\
-Other health metrics/engagement numbers (mainly show they are stable) \\
-Discuss flights on MUID vs ANIDs, also flight on Infopane (cover some issues
involved) \\ 
---For MUID vs ANID, control was ANIDs (for MUIDs no personalization),
experiment is both get the single CB model \\
---For Infopane, control is nothing, experiment is use CB on Infopane, again
single model for all \\
-how quickly does the model get good? \\
---mathematical requirements \\
---CTR plot increases at beginning of flight \\

\subsection{Implementation details and experimental setup}

The client library has been implemented in ...

- Describe experiments, also implementation details that are not fundamental to
our design. How much of this should be in a separate section (like here) vs
baked into section 4?

\subsection{Microbenchmarks}

\subsubsection{The \ds makes low latency decisions and learns from data quickly.}

- End-to-end latency \\
- Join server throughput (difference between alternative implementations) \\
- Options for scaling out training service
- Demonstration of failure recovery

\comment{Where should demonstration of real-time offline evaluation go?}

\subsection{\msn deployment}

\sscomment{
Working with \msn, we were able to benchmark some aspects of the live production
deployment. The Join Service based on Redis uses a tier of front-end servers to
group interaction data into $T$-second timeslices, by creating a (key, value)
pair for each $T$-second timeslice and storing a list of events in it. According
to the expiration duration of the application, a tier of front-end servers
periodically queries the Redis database to complete joined events. Thus the main
concern of \msn was whether each timeslice could be processed fast enough (at
least real-time).  We evaluated this using 6 front-end servers and a 3-node
Redis cluster during peak traffic time in the East US zone, with $T = 30$secs.
The average latency computed was \ldots
}
}

\ignore{ 
\sscomment{
Indeed, obtaining clearance for running experiments at \msn requires passing
performance regression tests that measure CPU/request, execution time, and
other metrics. The \ds increased these values by 4.9\% and 9.8\%, respectively.
Since \msn's rendering process is limited by CPU, the first metric is the most
important.  In addition to fast decisions, the \ds also enabled a low-latency
learning loop, which \msn leveraged to deploy models every 10 minutes.  Finally,
full reproducibility was also crucial in our debugging efforts leading up to the
Slate1 flight; we discuss this and other lessons in more detail next.
}
}

\section{Related work}
\label{sec:related}

Here we discuss other machine learning approaches related to \mwt,
including other experimentation and ML systems.

\subsection{Machine learning with exploration}
\label{sec:related-ML}

There are hundreds of papers related to exploration which broadly fall
into 3 categories. In \emph{active learning}~\cite{beygelzimer2010,
  hanneke2014, jamieson2015next, ICE}, the algorithm helps select
examples to label for a supervised learner. A maximally general
setting is \emph{reinforcement
  learning}~\cite{SuttonBartoRL-book98,CsabaRL-book-2010} where an
algorithm repeatedly chooses among actions and optimizes
\emph{long-term reward}. A simpler setting is multi-armed bandits
(MAB) where actions are chosen without contextual
information~\cite{Bubeck-survey12,Gittins-book11}. We build on
contextual bandits with policy sets
\cite{RegressorElim-aistats12,monster-icml14,bandits-exp3,policy_elim,Langford-nips07},
as well as offline policy evaluation
\cite{Dudik-uai12,DR11,Langford-www10,Langford-wsdm11}. The \mwt
capability enabled in this approach is typically absent from
alternatives such as Thompson Sampling~\cite{Thompson-1933}.

\ignore{Alternative
approaches for contextual bandits typically impose substantial
modeling assumptions such as linearity
\cite{Auer-focs00,Reyzin-aistats11-linear,Langford-www10}, Lipschitz
rewards \cite{Pal-Bandits-aistats10,contextualMAB-colt11}, or
availability of (correct) Bayesian priors \cite{Shipra-icml13}, which
often limits applicability.}

\subsection{Systems for ML and experimentation}
\label{sec:related-systems}

We previously discussed these systems with regards to technical debt
(\refsec{motivation}). A more general discussion follows.

\xhdr{A/B testing.} A/B testing refers to randomized experiments with
subjects randomly partitioned amongst treatments. It is routinely used
in medicine and social science, and has become standard in many
Internet services~\cite{KohaviLSH09,KohaviAB-2015}, as well supported
by statistical theory~\cite{FieldExptsBook-2012}. A more advanced
version, ``multi-variate testing", runs many A/B tests in
parallel. Several commercialized systems provide A/B
testing in web services (Google Analytics~\cite{GoogleAnalytics},
Optimizely~\cite{optimizely}, MixPanel~\cite{mixpanel}, etc.). The \ds
instead builds on \mwt, a paradigm exponentially more efficient in
data usage than A/B testing.

\xhdr{Bandit learning systems.}  \sid{Why isn't VW included here, and
  listed in supervised ML systems only?}  Several platforms support
bandit learning for web services. Google Analytics
\cite{GoogleAnalytics} supports Thompson Sampling
\cite{Thompson-1933}. Yelp MOE \cite{YelpMOE} is an open-source
software package which implements optimization over a large parameter
space via sequential A/B tests. Bayesian optimization is used to
compute parameters for the ``next" A/B test.\footnote{{\texttt
    SigOpt.com} is a commercial platform which builds on Yelp MOE.}
Clipper~\cite{clipper} uses bandit algorithms to adapt over supervised
ML systems. However, these systems do not support \emph{contextual}
bandit learning, and they do not instrument automatic deployment of
learned policies (and hence do not ``close the loop" in
Figure~\ref{fig:cycle}).

\xhdr{Contextual bandits deployments.}  There have been several
applications of contextual bandit learning in web services (\eg news
recommendation~\cite{Deepak,Langford-www10,Langford-wsdm11} and
advertising~\cite{bottou-ad}). However, they have all been one-offs
rather than a general-purpose system like the \ds.

\xhdr{Systems for supervised learning.} There are many systems
designed for supervised machine learning such as CNTK~\cite{CNTK},
GraphLab~\cite{GraphLab}, Parameter Server~\cite{parameterserver},
MLlib~\cite{MLlib}, TensorFlow~\cite{TensorFlow,TensorFlow-serve},
Torch~\cite{Torch}, Minerva~\cite{minerva} and Vowpal Wabbit~\cite{VW} to name a few.
These principally support Machine Learning model development.  A few
more, such as Google Cloud ML~\cite{GoogleML}, Amazon
ML~\cite{AmazonML}, and AzureML~\cite{AzureML} are designed to support
development and \emph{deployment}.  However, these systems do not
support data gathering or exploration. \sid{On that note, mention Clipper,
LASER, and TensorFlow serving as supporting model/prediction serving.}

\ignore{\sid{Replace with discussion of Clipper, as Velox is basically defunct
	predecessor.  Move
	to bandits paragraph?} Velox~\cite{velox-cidr15} is an open-source system that
	supports model serving and batch training.  Velox can adapt models online to users by adjusting preference weights in the user profile. It collects data that can be used to
retrain models via Spark~\cite{Spark}.  Velox does not perform exploration.}

We know of two other systems that fully support data collection with
exploration, model development, and deployment: LUIS~\cite{LUIS}
(based on ICE~\cite{ICE}), and NEXT~\cite{jamieson2015next}. These
systems support \emph{active learning},\footnote{NEXT does have some
  support for bandit algorithms, but does not provide the \mwt
  capability or fully general contextual bandits.} and hence make
exploration decisions for labeling in the back-end (unlike the \ds
which makes decisions for customer-facing applications),
and do not provide \mwt capability.

\section{Conclusion}
\label{sec:future}

\sid{replace with ``Discussion'' section, address T5?}
\sid{Include plans to make DS fully multi-tenant, which means join service and
online learner on shared resources.}

We have presented the \ds, the first general-purpose service for contextual
learning. It supports the complete data lifecycle and
combines an ML methodology with careful system design to address many
common failure modes.  Going forward, our goal is to make the service
completely parameter free.  We also plan to use \mwt capability to
provide more sophisticated safeguards for production deployments.


\ignore{
 combined ML methodology and system design which
automates many of the burdensome tasks that data scientists face such
as gathering the right data and deploying in an appropriate manner.
Instead, a data scientist can focus on the core tasks of finding the right
features, representation, or reward.

The data lifecycle support also makes basic application of the
Decision Service feasible without a data scientist.  To assist in
lowering the barrier to entry, we  are exploring techniques based on expert
learning~\cite{experts} and hyperparameter search that may further automate
the process. Since the policy evaluation techniques can provide
accurate predictions of online performance, such automations are
guaranteed to be statistically sound. We are also focusing on making the decision service easy
to deploy and use because we believe this is key to goal of democratizing machine learning
for everyone.

The Decision Service can also naturally be extended to a greater
variety of problems, all of which can benefit from data lifecycle
support.  Plausible extensions might address advanced variants like
reinforcement and active learning, and simpler ones like supervised
learning.
}
\small
\bibliography{bib-abbrv,decision_service,bib-bandits,bib-AGT,bib-ML,bib-slivkins}
\bibliographystyle{abbrv}

\end{document}